# Joint Prediction Regions for time-series models

Eshant English

Submitted for the Degree of Master of Science in

## MSc Machine Learning

Under the Supervision of Dr Nicola Paoletti

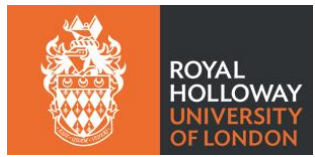

Department of Computer Science
Royal Holloway University of London
Egham, Surrey TW20 0EX, UK

Sept 5, 2021

# Declaration

This report has been prepared on the basis of my own work. Where other published and unpublished source materials have been used, these have been acknowledged.

**Word Count**: 16,118 (excluding Appendix)

**Student Name**: Eshant English

**Date of Submission**: Sept 5, 2021

**Signature**: eshant

# Abstract

Machine Learning algorithms are notorious for providing point predictions but not prediction intervals. There are many applications where one requires confidence in predictions and prediction intervals. Stringing together, these intervals give rise to joint prediction regions with desired significance level. It is an easy task to compute Joint Prediction regions (JPR) when the data is IID. However, the task becomes overly difficult when JPR is needed for time series because of dependence between the observations. This project aims to implement Wolf and Wunderli's method[1] for constructing JPRs and compare it with other methods (e.g. NP heuristic, Joint Marginals). The method under study is based on bootstrapping is applied on different datasets (Min Temp, Sunspots), using different predictors (e.g. ARIMA and LSTM). One challenge of applying the method under study is to derive prediction standard errors for models, it cannot be obtained analytically. A novel method to estimate prediction standard error for different predictors is also devised. Finally, the method is applied on a synthetic dataset to find empirical averages and empirical widths and the results from the Wolf and Wunderli paper are consolidated. The experimental results show narrowing of width with strong predictors like neural nets, widening of width with increasing forecast horizon $H$ and decreasing significance level α, controlling the width with parameter $k$ in $k - FWE$, loss of information using Joint Marginals.

# Contents

# 1 Introduction

A Machine Learning prediction problem is a prediction problem of a random variable; due to its random nature, it is often insufficient to talk only about point forecast. This insufficiency is dealt with by talking about prediction intervals, which provides a statement about the point estimate. These prediction intervals for prediction are equivalent to confidence intervals for population parameters. Whilst the task of computing confidence intervals is easily achievable, the computation of prediction intervals is inherently difficult in comparison.

When dealing with time series, the prediction task is about forecasting at some point in future. We don't always require point forecasts when dealing with time series; we also desire the most likely forecast sequence up to some point $h$ in future. In that case, the statement around its uncertainty is provided by a joint prediction region. In other words, a Joint Prediction Region for time series is the region that contains the sequence of predictions with a desired significance level $\alpha$. The task of constructing a joint prediction region is often ignored in the literature.

A simple way of computing the Joint Prediction region is to string together marginal prediction intervals with significance $\alpha$, but due to dependency on previous observations, such a region has a small volume and fails to provide desired coverage of $1-\alpha$. One quick solution is to apply a correction for multiple independent tests, e.g. Bonferroni correction, which results in a region with high volume and coverage higher than $1-\alpha$ due to loss of information. It's this dependency which we need to exploit to find joint prediction regions with the desired coverage. This task is not trivial and is much more difficult than computing prediction intervals. One important advancement in the direction is the JPR by Wolf et al.

The main aim of this dissertation is to peruse and implement the Joint Prediction Region paper from Wolf et al. [1] and compute empirical coverage and width for a chosen dataset. The empirical coverages and widths are computed for different values of significance level $\alpha$, horizon length $H$, and other hyperparameters to explore the impact of these hyperparameters. As extensions, this dissertation also compares different JPR methods and performance of the JPR by Wolf et al. on different predictors and different datasets.

This project has provided me with a deep understanding of concepts like bootstrapping, different time-series models. It has given me exposure to the research environment and papers and prepared me to read high-level journals. It has strengthened my understanding of deep learning. It has also motivated me to work in the area of uncertainty, which has now become my research interest. It was thanks to this project I was able to decide my future career, as I thrive to become a leading researcher in the field of Machine Learning.

# 2   Background Research

This chapter is mostly adapted from Advanced Data Analysis from Elementary point of view [2], Dive into Deep Learning[3], Forecasting Principles and Practice [6], and Time Series Analysis and its applications[7].

## 2.1   Time Series

Traditionally Machine Learning algorithms require IID (Independent and Identically Distributed) assumptions before they can be applied. However, not always the data points are independent of each other. There are many applications where the produced data points are dependent on each other. The simplest form of dependent data is time series. As the name suggests, they are series of data/observations recorded over time. Time series are typically recorded over equally spaced time intervals starting from t, denoted as $X_t, X_{t+1}, X_{t+2}, \ldots$ or $X(t), X(t+1), X(t+2)$ and so on. Time series can be viewed as a realisation of some underlying partially random or stochastic process generating the data. This data is used to make inferences about the process and make predictions about the future. As each observation in the series has a dependency on previous observations, the task of prediction becomes more difficult than IID data. The aim is to build models to find the structure of dependence and exploit it to make inferences about future observations; these models are dubbed as Time Series models.

The observations in a time series are mostly sampled at regular time intervals, also known as the sampling rate. There are also time series sampled with irregular intervals ($t_i - t_{i-1} \neq t_{i+1} - t_i$) which makes them difficult to deal with.

### 2.1.1   Property of Stationarity

In independent data, the property of the same distribution for all the data points had made prediction possible, and similar property for dependent data is a desirable property to have. For time series, this property is called stationarity, which implies whilst the value of observations in a time series can be different, but their distribution should be the same.

**Definition 2.1 (Strong Stationarity)** Formally, a time series is considered strictly stationary if $X_{1:k}$ and $X_{t:t+k}$ have the same distribution for all k and t. The distribution of blocks of length k is time-invariant.

To reiterate, it does not imply each block of length k has the exact string of values, but the distribution is the same for each block. Generally, it is hard to have a strictly stationary time series, but for most statistical applications, it is enough to have what is called weak stationarity.

**Definition 2.2 (Weak Stationarity)** For a time series to be weakly stationary, every observation in the time series must have the same mean, and any pair of observations must have the same covariance. $\mathbb{E}[X_1] = \mathbb{E}[X_t]$ $and$ $Cov[X_1, X_k] = Cov[X_t, X_{t+k-1}]$ .

A strict/strong stationarity in time series always implies weak stationarity, but not vice versa. (An exception is when the time series follows a Gaussian process). It must be noted that whilst strong stationarity deals with blocks of time, weak stationarity does not. An important mention about time series is the ergodic theorem, which is the basis of statistical inferences for time series data. It is the time series counterpart of the law of large numbers.

**Definition 2.3 (Ergodic Theorem)** Ergodic theorem states that the time average converges on expectations only for $X_t$ itself. In other words, a single long time series becomes representative of the whole data generating process[2]. In the simplest mathematical form,
$$\overline{X_n} \to \mathbb{E}[X_1]$$
Where $\overline{X_n}$ is the time average and $\mathbb{E}[X_1]$ is the expectation of $X_1$.

## 2.2 Time Series Models

Many time series models exist, many of them are a counterpart of the models for IID data. However, using some of these models (e.g. ARMA) requires making sure the time series is stationary (weak or strong). If it's not stationary, it is required to convert a non-stationary time series into a stationary time series. It can be done by transforming the time series; one of the most successful ways is differencing, which removes the trend present in a time series.

Traditionally, the most popular time series models are ARMA and ARIMA models, Markov and Hidden Markov models, Exponential Time Smoothing models, etc. Recent time series models include Vector Autoregression models, Prophet models, Bootstrapping and Bagging models, Neural Network models. All these models have their pros and cons. For this section, Exponential Smoothing ARIMA models and LSTM neural networks are discussed in detail as they are the main models used for analysis and prediction.

### 2.2.1 Exponential Smoothing Models

Exponential smoothing is amongst the oldest methods for forecasting. The predictions produced by Exponential Smoothing are weighted averages of past observations—these weights decay exponentially, which gives the name exponential smoothing. The theory behind the method is very simple, the recent observations are given a higher weight, and the less recent observations are given less weight. Due to its simplicity, it provides forecasts very quickly. Exponential smoothing has several variants: Simple exponential smoothing, used when there is no trend or seasonal component in a series; Holt's method, which incorporates trend of a series; and Holt-Winter method, which incorporates seasonal component in the series.

The equation modifies to include trend and seasonality with additional smoothing parameters β and γ, respectively.

**Simple Exponential Smoothing**

It is Exponential Smoothing in its primitive form. It does not take into account the trend or seasonality of a time series before making a prediction.

Simple Exponential smoothing can be written mathematically as

$$\hat{y}_{T+1|T} = \alpha y_T + \alpha(1-\alpha)y_{T-1} + \alpha(1-\alpha)^2 y_{T-2} + \cdots$$

Where α is the smoothing parameter, i.e. $0 \leq \alpha \leq 1$ and $\hat{y}_{T+1|T}$ the prediction at time $T + 1$, conditional on past $T$ observations.

Exponential Smoothing can also be written as a weighted average of the most recent observation and $y_T$ and the previous forecast $\hat{y}_{T|T-1}$

$$\hat{y}_{T+1|T} = \alpha y_T + (1-\alpha)\hat{y}_{T|T-1}$$

Where $\hat{y}_{T|T-1}$ the forecast at time $T$ conditional on past $T - 1$ observations.

An alternative representation of Simple Exponential Smoothing is in component form as a time series is divided into three components, trend, seasonality and level. It is convenient to view the application of Simple Exponential Smoothing as finding the level of a series. This form helps when comparing the variants of Exponential Smoothing. The component form consists of two equations one is for forecasting, and another is for smoothing.

Forecasting component: $\hat{y}_{t+h} = l_t$
Smoothing component: $l_t = \alpha y_t + (1-\alpha)l_{t-1}$

where $l_t$ is the level of the series at time $t$. The previous equation weighted can be recovered from this equation by simply replacing $l_t$ by $\hat{y}_{t+1}$ and $l_{t-1}$ by $\hat{y}_t$.

One problem with Simple exponential smoothing is that it produces flat forecasts, which means the forecast remains the same for any value of $h$. This can be seen by the forecasting component of the above equation.

**Double Exponential Smoothing**

Double Exponential Smoothing or Holt's method extends Simple Exponential Smoothing by incorporating a trend component. Doing this increases the smoothing equation form from one in simple exponential smoothing to two. The component form equations are

Forecasting component: $\hat{y}_{t+h} = l_t + hbt$
Level Smoothing component: $l_t = \alpha y_t + (1-\alpha)(l_{t-1} + b_{t-1})$
Trend Smoothing component: $b_t = \beta(l_t + l_{t-1}) + (1-\beta)b_{t-1}$

where $\beta$ is trend smoothing parameter, and $b_t$ accounts for trend at time $t$

The trend component ensures the forecast is not flat. However, it is not always desirable for the longer forecast to use the same trend. There is no guarantee that the trend will keep continuing for longer forecasts. This motivated a variant of Holt's method, which dampens the trend when forecast interval $h$ grows larger. Using this variant causes the forecast to flat out at some point. Also, the equations shown here assume the trend is additive; some versions modify it to include multiplicative trends.

**Triple Exponential Smoothing**

Triple Exponential Smoothing or Holt-Winters' method adds another improvement to Exponential smoothing. It adds the seasonal component, which creates a third smoothing equation. The resulting component form is

Forecasting component: $\hat{y}_{t+h} = l_t + hbt + s_{t+h-m(k+1)}$
Level Smoothing component: $l_t = \alpha(y_t - s_{t-m}) + (1-\alpha)(l_{t-1} + b_{t-1})$
Trend Smoothing component: $b_t = \beta(l_t + l_{t-1}) + (1-\beta)b_{t-1}$
Season Smoothing component: $s_t = \gamma(y_t - l_{t-1} - b_{t-1}) + (1-\gamma)s_{t-m}$

where $\gamma$ is the seasonal smoothing parameter, $k$ is the integer part of $(h-1)/m$, which ensures the estimation of seasonal indices come from the recent season, and $m$ is the seasonality of the series.

Like in double smoothing, these equations consider additive seasonality. There are some variations of the method, which includes multiplicative seasonality. There can be many variant methods based on different combinations; there can be a damped trend with multiplicative seasonality or a multiplicative trend with additive seasonality.

### 2.2.2 ARMA and ARIMA models

A time series can be decomposed into three terms: Trend, Seasonality, and an irregular Error term with no pattern. A stationary series has no obvious pattern in it. Therefore, to apply ARMA models, it is required to remove any trend or seasonality present in the series to make it stationary. It is to be noted that a stationary series may have a cyclic behaviour, but it should not be seasonal, i.e. have a fixed period. As mentioned before, trend and seasonality can be removed by using differencing. It is usually applied to the time series by differencing the consecutive observations for removing trend or differencing observations with seasonality $m$. A time series can be differenced multiple times until it becomes stationary. One statistical technique to determine stationarity in series is by applying the unit root test. Sometimes, it is required to transformations other than differencing to make the series stationary, and one such example is the Box-Cox transformation.

**Autoregressive models**

In a typical linear regression model, we use a linear combination of predictors to forecast the random variable of interest/ dependent variable. Autoregressive models are the counterpart of linear regression for time series data. In Autoregressive models, the future value of a variable (observation of a time series) is forecasted (or regressed) over past values of the time series. The order of Autoregressive models, denoted by $p$, is the number of past observations used to forecast future observations. AR models are very flexible at handling a wide range of different time series.

The model can be written as:

$$y_t = c + \phi_1 y_{t-1} + \phi_2 y_{t-2} + \cdots + \phi_p y_{t-p} + \varepsilon_t$$

Where $\varepsilon_t$ is white noise and lagged values of $y_t$ are the predictors, and $c$ can be seen as a constant in a regression model. This model is referred to as **AR(p)** model. The value of $p$ is typically obtained by looking at the series' autocorrelation function (ACF) or partial autocorrelation function (PACF). Typically, ACF tails off gradually for pure AR models, but the PACF cuts off at lag $p + 1$.

**Definition 2.4 (Autocorrelation Function)** Autocorrelation function measures the extent of linear relationship between the lagged values of a time series. It is defined as

$$r(k) = \frac{\sum_{t=k+1}^{n}(x_t - \overline{x})}{\sum_{t=1}^{n}(x_t - \overline{x})^2}$$

where $r(k)$ is the correlation coefficient with lag $k$, $x_t$ is the time series and $\overline{x}$ is the mean of the series, and $n$ is the total number of observations.

**Definition 2.5 (Partial Autocorrelation Function)** Partial Autocorrelation function (PACF) measures the relationship between $x_t$ and $x_{t-k}$ after removing the effects of Lag $1,2,3,\ldots,k-1$. Formally

$$\alpha(k) = r(k) - (effect\ from\ intermediate\ lagged\ values)$$

where $\alpha(k)$ is the partial correlation coefficient with lag $k$.

**Moving Average models**

Moving Average models use a similar concept as Autoregressive models. Still, instead of using the past values of the variable, they use past forecast errors in a regression-like model. The order of Moving Average (MA) models, denoted by $q$, is the number of past forecast errors used to forecast the future observations. Since we do not observe the values of $\varepsilon_t$, it is not a regression in the usual way.

The model can be written as

$$y_t = c + \varepsilon_t + \theta_1 \varepsilon_{t-1} + \theta_2 \varepsilon_{t-2} + \cdots + \theta_q \varepsilon_{t-q}$$

where $\varepsilon_t$ is white noise, $\theta_i$ are model parameters for $\varepsilon_{t-i}$ where $i$ ranges from 1 to $q$, and $c$ can be seen as a constant in a regression model. This model is referred to as **MA(q)** model. Each value $y_t$ can be viewed as a weighted moving average of the past few forecast errors. However, the coefficient may not necessarily sum to 1. Like AR models, the order $q$ can be obtained by looking at the series' autocorrelation and partial autocorrelation functions. Typically, PACF tails off gradually for pure MA models, but the ACF cuts off at lag $q + 1$.

An interesting property of that an $MA(q)$ model can be written as an $AR(\infty)$ model after some mathematical manipulation. Similarly, it is possible to write an $AR(p)$ model as $MA(\infty)$ model.

**ARMA models**

ARMA models are obtained after we combine AR and MA models. An $ARMA(p, q)$ model has $AR(p)$ and $MA(q)$ parts. It can be written as:

$$y_t = c + \phi_1 y_{t-1} + \cdots + \phi_p y_{t-p} + \theta_1 \varepsilon_{t-1} + \cdots + \theta_q \varepsilon_{t-q} + \varepsilon_t$$

As it can be seen, it has all the components from the AR and MA models.

**ARIMA models**

ARIMA models are more general than ARMA models; whilst ARMA models require the time series to be stationary, ARIMA models can work with non-stationary series. The "$I$" in ARIMA stands for integrated. They can be denoted as $ARIMA(p, d, q)$, where $d$ is the differencing operation. The time series can be differenced as many times as necessary to make it stationary. However, one must be cautious not to overdo the differencing as it could introduce correlations and patterns not present in the original time series. It can be written as:

$$y'_t = c + \phi_1 y'_{t-1} + \cdots + \phi_p y'_{t-p} + \theta_1 \varepsilon_{t-1} + \cdots + \theta_q \varepsilon_{t-q} + \varepsilon_t$$

Where $y'_t$ is the differenced version of the original time series $y_t$.

Both PACF and ACF gradually tails off for ARMA and ARIMA models, making it harder to find the best values of $p, q$ to model the time series. One suitable method to find $p, q$ for ARMA and ARIMA models is to use Akaike's Information Criteria (AIC) or Bayesian Information Criterion (BIC). AIC or Akaike's Information Criteria and BIC or Bayesian Information Criterion are two popular methods for model selections. The model with minimum AIC or BIC is selected. These methods, whilst helpful in finding $p, q$ are inept at finding $d$ for ARIMA models. An approach might be to find appropriate differencing by visualisations and unit root test, before applying them to find $p, q$.

**Definition 2.6 (AIC)** AIC or Akaike's Information Criteria is a method used for model selection. It uses a measure of goodness of fit to find the best model. AIC is defined as
$$AIC = 2k - 2ln(\hat{L})$$

where $k$ is the number of estimated parameters in the model, and $\hat{L}$ is the maximum value of likelihood function of the model.

**Definition 2.7 (BIC)** BIC or Bayesian Information Criterion is a method used for model selection. It is based partly on the likelihood function and uses a parametric approach to find the best model. BIC is defined as
$$BIC = kln(n) - 2ln(\hat{L})$$

where $k$ is the number of estimated parameters in the model, $n$ is the number of observations, and $\hat{L}$ is the maximum value of likelihood function of the model.

**Seasonal ARIMA models**

Seasonal ARIMA models are a modification of ARIMA models, which includes seasonality. They can be viewed as having two components one is non-seasonal, and the other is seasonal. The non-seasonal component can be represented as $(p, d, q)$ and the seasonal component can be represented as $(P, D, Q)_m$ where $m$ is the seasonality.

**Choosing the best ARMA/ARIMA model**

It is usually considered good practice to fit a model with fewer parameters, as more parameters can often lead to overfitting. Choosing between different orders of ARMA/ ARIMA models can be a difficult task. One approach is to see the residual errors after fitting an ARMA/ARIMA model; if the residuals appear to be white noise, the application of the ARIMA model can be considered successful. Otherwise, choosing a different model will be a better option. A statistical method to use for finding if the residuals are white noise is the portmanteau test for residuals. We would be using the same methodology whilst choosing the appropriate model for the chosen case studies.

**Definition 2.8 (Portmanteau test)** Portmanteau test determines if the first $l$ autocorrelations are significantly different from a white noise process or not. Two examples for the Portmanteau test are Box-Pierce test[19] and Ljung-Box test[20].

$$\text{Ljung-Box}(k) = n \times (n+2) \times \sum_{k=1}^{K} \frac{r_{e,k}}{n-k}$$

$$\text{Box-Pierce}(k) = n \sum_{k=1}^{K} r_{e,k}$$

where $r_{a,k}$ is the autocorrelation coefficient of the residuals $e$ at lag $k$, $K$ is the maximum lag considered, and $n$ is the number of terms in differenced series.

### 2.2.3 Artificial Neural Net

Artificial Neural Networks, also referred to as Neural Nets, are one of the most powerful machine learning models. They allow complex non-linear relationships between the response variable and the predictor variables. As explained at the beginning of the section, most models for dependent data have either time-series counterparts or can be adapted for a time series. More powerful neural networks have more layers allowing more flexibility and better predictions. The learning in neural networks minimises the "cost function" by assigning weights to neurons in a neural net. These networks use differentiation to find the gradient of the cost function with respect to weights and thus use gradient descent to find the minimum of the "cost function". One disadvantage of neural networks is that they are slow to learn and require a lot of data. Whilst Basic neural networks can be used for time series, they don't provide a good forecast because they don't use the temporal information of the time series. For one case study, we used Neural network autoregression for forecasting by using one hidden layer. It makes one forecast at a time and uses that forecast to predict the forecast at the second step, iteratively.

There are other variants of Neural Networks, including Convolutional Neural Networks, Recurrent Neural Networks. Both can be used for time series forecasting. However, Recurrent Neural Networks are of higher significance as they are capable of using temporal information. They are sequential models and therefore are slower than other types of neural networks. They suffer from a problem called Gradient Vanishing; in simple terms, it means that these models are incapable of using information from the distant past (maybe after ten observations). It is analogous to how we humans remember our recent experiences more than those in the past. This problem can be mitigated by Attention Models, GRU (Gated Recurrent Units), and LSTM (Long Short-Term Memory) models. We used the LSTM model in our case studies and will discuss it in greater detail. Before discussing LSTM, we want to mention Transformer models, which uses the attention mechanism used in Attention models but removes the sequential learning process, thus significantly improving learning time. Transformer models are one of the fastest and most powerful models for sequential data in recent times.

### 2.2.4 LSTM

LSTM (Long short-term memory) models are one of the most successful models for time series. They can be considered a recurrent neural network and a more complicated version of GRU (Gated Recurrent Units) models. LSTM's design is motivated by logic gates in a computer system. The main idea behind LSTM's is to have a memory cell that holds the past information and some logic gates that manipulate the cell's state. These gates decide how much impact the information from a cell would have on the next step's hidden state, how much new information should be added to the memory cell, deciding what information should be discarded, e.t.c. There are five essential components of an LSTM that helps in producing a forecast.

- Cell State ($c_t$)- It represents the internal memory of a cell at time step $t$, it stores both short-term and long-term memories.
- Hidden State ($h_t$)- This is the hidden/output state at every time step $t$, it is calculated by the input, the previous hidden state $h_{t-1}$, and current cell value. It can decide to retrieve short-term or long-term, or both types of memory from the cell state.
- Input gate ($i_t$)- This gate decides the flow of information from the current input that should be added to the cell state.
- Forget gate ($f_t$)- This gate decides the flow of information from the current input and the previous cell state that should be retained in the current cell state.
- Output gate ($o_t$)This gate decides how much information from the current cell state flows into the hidden state, enabling LSTM to pick only long-term or short-term memories as necessary.

The equations for calculating each of these components are below:

$$i_t = \sigma(W_{ix}x_t + W_{ih}h_{t-1} + b_i)$$
$$\tilde{c}_t = \sigma(W_{cx}x_t + W_{ch}h_{t-1} + b_c)$$
$$f_t = \sigma(W_{fx}x_t + W_{fh}h_{t-1} + b_f)$$
$$c_t = f_t c_{t-1} + i_t \tilde{c}_t$$
$$o_t = \sigma(W_{ox}x_t + W_{oh}h_{t-1} + b_o)$$
$$h_t = o_t tanh(c_t)$$

Here, $W$ refer to weight matrices, and $\sigma$ refers to sigmoid function. The figure below shows the internal representation of an LSTM cell.

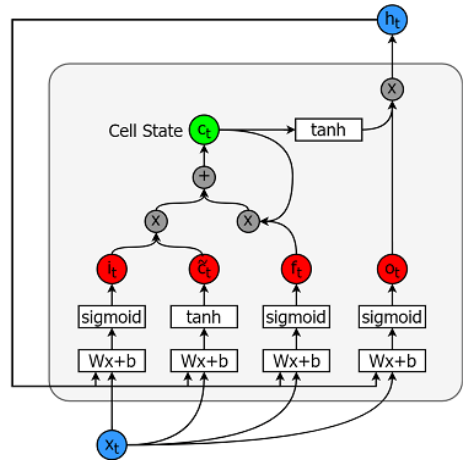
**Figure 1:** LSTM cell [4]

## 2.3 Bootstrapping

Statistics is the branch of science dealing with limited and imperfect data to draw inferences about the population. The inferences drawn from the available data is imperfect, and it would be a mistake to consider it certain. This can be remedied by finding uncertainty around the inferences drawn from the data.

Mainly, the quantities drawn from the data can be represented as the function of the underlying probability distribution. These quantities can be called Functionals or Statistical Functionals or sample statistics as they are a function of a function. Since rerunning the same experiment under the same condition causes variability in results, the functionals/inferences obtained also exhibit variability. One way to account for this variability is to calculate a confidence region/interval, which could provide some insight into the values producing the result with some probability. A sampling distribution is required to calculate such intervals, which is not always possible to obtain and can be very expensive to obtain as well. Bootstrap Sampling by Bradley Efron[17] provides a convenient and inexpensive way to obtain a sampling distribution from a single sample.

**Definition 2.9 (Sampling Distribution)** Sampling distribution refers to the probability distribution of a statistic obtained from a sample drawn from a certain population.

The idea behind bootstrap sampling is to simulate replication from a given set of data. Bootstrap sampling can be done in two major forms, i.e., model-based bootstrap and model-free bootstrap.

### 2.3.1 Model-based Bootstrap Sampling

The parametric model-based bootstrap sampling requires fitting a model to the given data, which acts as a guess at the data generating mechanism. The generated mechanism is then used to simulate the data, which, ideally, should have the same distribution as the original data. Feeding the simulated data through our estimator gives us one draw from the sampling distribution, and its repetition gives rise to the sampling distribution, which can be used to find standard error or confidence interval. In a way, the model accounts for its own uncertainty—a Schematic for model-based bootstrap is shown in the figure below.

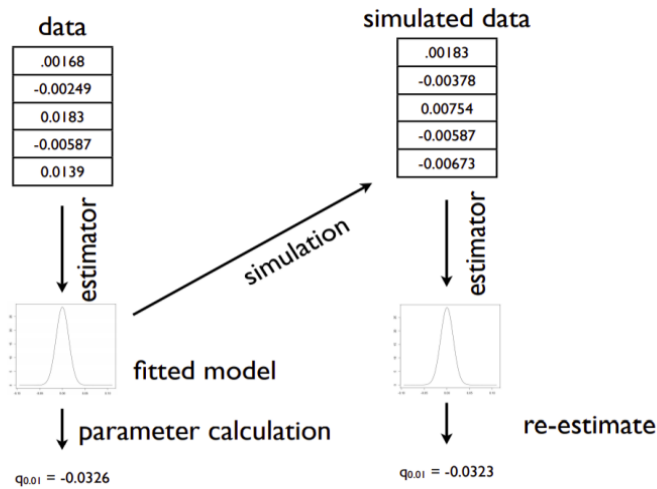

**Figure 2**: Model-Based Bootstrapping[2]

The key to effective or successful bootstrap sampling is to guess the data generating mechanism accurately. If the fitted model is inaccurate, the inferences obtained from it would be misleading.

As bootstrap approximates the sampling distribution, it can have three sources of approximation errors.

1) Simulation Error- Caused by using finitely many replications to stand for the full sampling distribution. It can be made relatively very small by using brute force, i.e. using just enough replications.

2) Statistical Error- Caused if the sampling distribution obtained by bootstrapping is not exactly the same as the sampling distribution under the true data generating mechanism. It is generally solved by bias correction.

3) Specification Error- Caused when the data source does not follow our model at all. In this case, simulating the model never matches the sampling distribution. A clever way to reduce it is to use Model-Free Bootstrap Sampling described in the next section.

### 2.3.2 Model-free Bootstrap Sampling

An alternative to the traditional parameterised model-based bootstrap Sampling is to do it without assuming a model. In order to do this, the data is treated as the whole population, and the surrogate data/simulated data is drawn from the original data with replacement. It is also called Resampling Bootstrap. Generally, It's the bootstrap method meant when the term 'bootstrap' is used without a modifier. The reason this method works well is that the empirical distribution is the most unbiased. Therefore, the only thing changing from the Model-Based method is the distribution from which the surrogate data is obtained.

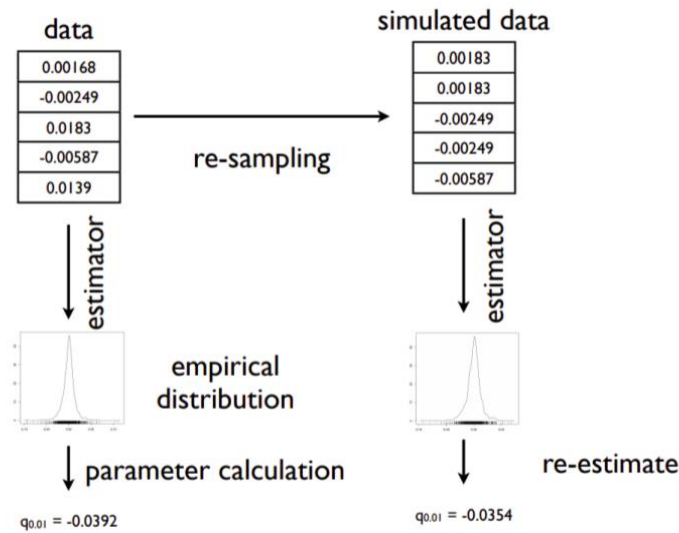

**Figure 3:** Model-Free Bootstrapped Sampling[2]

A variant of this is called Smoothed Bootstrap, in which some Gaussian noise is added after resampling the data points.

### 2.3.3 Model-free Bootstrap Sampling vs Model-based bootstrap sampling

Both the methods have their own merits. Model-Based Bootstrap Sampling generally leads to faster convergence and leads to more accurate results if the model is correctly specified. But it can give poor results if the model is misspecified, which can cause convergence to a wrong distribution. On the other hand, Model-Free Bootstrap Sampling has slower convergence and could provide larger intervals but doesn't require any model. A safe approach would be to use Model-based Bootstrapping only when there is high confidence in the model's correctness and use Model-free Bootstrapping otherwise.

### 2.3.4 Bootstrapping for Time Series

The Bootstrapping discussed so far is ideal if the data points/vectors are independent and identically distributed. However, if there is a dependence between them, the resampling cannot be done as usual. This demands specific changes to our process to incorporate these dependencies in our surrogate/simulated data. A critical type of dependent data is data in the form of a time series. It requires certain tricks to apply bootstrapping to it. Just like the normal bootstrapping methods, it can be divided into model-based and model-free methods.

The model-based bootstrapping for time-series is conceptually very similar to non-time-series bootstrapping. We require fitting a generative model which could be used to generate a full time series. We can use ARMA/ARIMA models to fit a model and can bootstrap from it. However, confidence in our fitted model is the only issue to achieve. If we are unsure about the fitted model, then we have to use model-free methods. Some of them are described below:

**Block Bootstraps**

Since simple resampling is useless because it destroys the dependencies between successive values in a time series, a clever trick by Kunsch (1989) preserves the dependencies and is simple to implement and understand. The idea is to take the full time series $X$ of $n$ elements and cut it into overlapping blocks of length $k$, such that blocks are $X[1:k], X[2:k+1], X[3:k+2]$ and so on down to $X[n-k+1:n]$. Then, draw $m = n/k$ of these blocks with replacement and put them in order. This will be one bootstrap series.

Doing this has preserved dependence between observations within each block. The resulting bootstrapped time series still has independence between blocks but is better than simple resampling of observations which can be seen as $k = 1$. Growing the value of $k$ can reduce this independence. The optimal number of folds to preserve properties of a times series can be shown to be $k = O(n^{1/3})$[11]. Block Bootstrapping has inspired several variants.

**Circular Bootstraps**

The idea behind Circular Bootstraps is to wrap the entire time series in a circular fashion and then sample $n$ blocks of length $k$. This has the advantage of better using the information we have about the dependence of distances $< k$.

**Block of Blocks Bootstraps**

This consists of first dividing the series into blocks of length $K1$ and then dividing each block into subblocks of length $K2$ $as$ $(K2 < K1)$. To generate a new series, we first sample blocks with replacement and then sample subblocks within each block with replacement. This preserves longer range dependencies at the cost of one extra parameter, $K2$.

**Stationary Bootstraps**

In this bootstrap method, the length of each block is random, chosen from a geometric distribution of $k$. Once the random sequence of $k$ is obtained, the blocks can be sampled with replacement. The basic block bootstrap doesn't quite provide us with stationary time series, and the distribution gets strange near the boundaries. Using this method solves those issues. It is useful when there is a strict requirement for a series to be stationary. However, the stationary bootstrap has slower convergence than the above bootstraps.

**Sieve Bootstraps**

It is a unique method that provides a compromise between model-based and model-free methods. This also requires fitting a model, but it does not fully believe in them. Instead, it finds a reasonable model and grows its complexity as more data is acquired. One popular choice is to use $AR(p)$ models, which are easy to fit.

**Bootstrapping for non-stationary Time Series**

The methods discussed above for bootstrapping a time series mostly work well when the series is stationary (unless model-based bootstrapping is applied). However, some changes need to be made if bootstrapping is required for a non-stationary time series. As discussed in the Time Series section, a non-stationary series can be made stationary by detrending or de-seasoning (removing the seasonal component). The resulting series can be used with any of the discussed bootstrapping methods, and then the seasonal component (if present) and the trend can be added to it to get a bootstrapped time series. For the case studies, we have preferred to use STL decomposition of the time series before bootstrapping, which provides several advantages over classical decomposition.

**Definition 2.10 (STL decomposition):** STL stands for "Seasonal and Trend Decomposition using Loess", where Loess is a method used to estimate non-linear relationships. STL is considered a versatile and robust method with more control on decomposition.

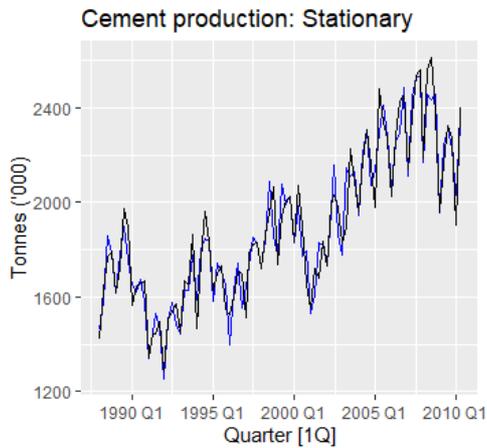 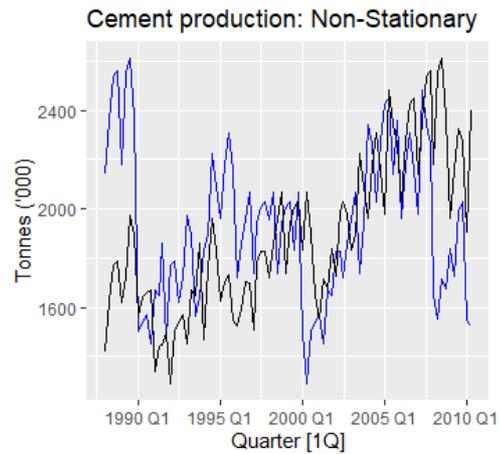

**Figure 4(a)** Bootstrapping on Stationary Series  **Figure 4(b)** Bootstrapping on non-stationary series

The figures above show bootstrapped versions in blue and real series in black. The figure on the left is obtained by first changing the series into stationary and then bootstrapping and adding back seasonal and trend components. The figure on the left shows bootstrapping the non-stationary series directly. As expected, the figure on the left provides a good surrogate series, whereas the series on the right does not. The above graphs are built-in $R$.

### 2.3.5 Final words on Bootstrapping

The main principle is that the sampling distribution generated under the true process should be close to the sampling under good estimates of the truth. If minor changes to the data generating process cause huge variations in the sampling distribution, bootstrapping will fail spectacularly. This means that small changes to parameters for Model-based bootstrapping should not make huge changes to the functionals. For model-free bootstrapping, adding or removing a few data points must not significantly change functionals of interest.

# 3 Joint Prediction Regions

Most machine-learning predictions provide a point prediction for a random variable of interest. In many cases, a point prediction is of little interest, and a prediction interval is desired, which accounts for uncertainty in that prediction. Many times, rather than the forecast of a random variable of interest for a single moment in time, it is desired to obtain the forecast over several periods into the future. The sequence of individual forecasts, one period at a time, is called Path forecast and the sequence of individual prediction intervals over several periods into time constitutes a prediction region. This region accounts for the uncertainty over several periods and has a high probability of finding the sequence of the random variable. In other words, A joint prediction region is a prediction region associated with our forecasting with a specified probability of finding the true sequence. The problem of finding the joint prediction region with the desired coverage probability $1 - \alpha$ is rather neglected in literature so far.

The material in this section has been adapted from Journal of Time Series Analysis J. Time. Ser. Anal. 36: 352–376 (2015).[1]

## 3.1 Confidence Intervals and Prediction Intervals

The situation of finding a prediction interval is like that of a confidence interval. Both are intervals to define uncertainty in our random variable of interest. The terminology '**Confidence Interval**' is used when uncertainty in a point estimate of a population parameter is calculated. In contrast, the **prediction interval** is used when that uncertainty is measured around a point prediction. However, the problem of constructing a prediction interval for a variable of interest is inherently a lot more complex than that of constructing a confidence interval for a variable of interest.

**Definition 3.1 (Confidence Interval and Prediction Interval)**

The Confidence Interval (CI) and Prediction Interval (PI) can be defined as

$$CI: \boldsymbol{P}\big(\theta \in C(X_1, \dots, X_n)\big) = 1 - \alpha$$

$$PI : \boldsymbol{P}(Y_i \in C(X_i)) = 1 - \alpha$$

where $\alpha$ is the significance level, $C$ is the interval, $\theta$ is the true parameter for which CI is defined, $Y_i$ is the true prediction for which PI is defined, $\boldsymbol{P}$ is the probability. CI and PI are defined as the interval $C$ which has probability $1 - \alpha$ for finding the true parameter or prediction. For CI interval $C$ is a function of the sample $(X_1, \dots, X_n)$; For PI interval $C$ is a function of observation $X_i$ for which prediction is desired.

In order to construct a confidence interval, the central limit theorem can be applied to show that the distribution of the estimated parameter is normal or approximately normal for large sample sizes. This allows the construction of standard, normal-theory confidence intervals. This can be used to find standard error for the parameter and find uncertainty at a desired significant level α. Alas, the central limit theorem cannot be used to show that the difference between a point forecast and the random variable of interest is approximately normal for large sample sizes. This difference is not normally distributed in practice, and assuming it to be normal would be a strong assumption likely to be violated. The use of bootstrapping to form intervals is motivated by higher-order considerations, namely when normal theory fails. Bootstrapping enables the formation of asymptotically consistent prediction intervals. The key problem is applying the bootstrap to construct asymptotically consistent prediction intervals with finite sample properties. A solution to this problem is provided by De Gooijer and Hyndman (2006, Section 12).[12]

**Definition 3.2 (Finite Sample Property)** Finite sample property means that a statistical method (e.g., for prediction intervals) produces correct results despite finite/small samples and not just asymptotically.

## 3.2 Constructing Joint Prediction Region (JPR)

Constructing a prediction interval for any period $h$ in $H$ (where $H$ is greater than or equal to $h$) with a desired significance level $\alpha$ is not difficult. It can be constructed using standard statistical methods. However, the task of constructing a joint prediction region is much more difficult. There are a few methods for constructing JPR: Marginal JPR, which joins Marginal Prediction Intervals; Scheffé joint regions, which inverts classical F-test; Jordà and Marcellino (2010)[8], which uses an asymptotic method but relies on a strong assumption of normal distribution of prediction error; Staszewska-Bystrova (2011)[9], which uses a bootstrap method of heuristic nature; and Wolf and Wunderli bootstrapped method.

The main aim method discussed in this report is from Wolf and Wunderli (2015), and we will have our main focus on it. It has been proven to contain the future path with a probability $1 - \alpha$, at least asymptotically, under mild high-level assumption.

### 3.2.1 Marginal JPR

One possible way to find a Joint Prediction Region would be to compute marginal prediction interval at the desired significance level for every period in $H$ and string together all the prediction intervals to get a joint prediction region for the sequence of $H$ periods. However, the JPR computed during this approach doesn't have the desired coverage probability of $1 - \alpha$. The coverage probability is, in fact, much lesser than $1 - \alpha$ and would be decreasing with increasing H. If we consider $E_h$, the event that the random variable at period $h$ in future will fall under its prediction interval, and consider all $E_h$ where $h \in [1, H]$ independent from each other, this JPR provides a probability of the future path to fall into the region equals $(1 - \alpha)^H$. But the assumption of independence is a little vague. In practice, there is always some dependence between the events, and the exact probability can only be obtained after knowing the dependency structure between the events. However, It can be said that JPR contains the future path with a probability ranging between $max(0, 1 - H \times \alpha)$ and $1 - \alpha$, where the lower bound is obtained using Bonferroni's inequality.

We can use Bonferroni's correction to find a conservative JPR by stringing together all the marginal prediction intervals at a significance level of $\alpha/H$ instead of $\alpha$. However, this method is crude and results in unnecessarily wide JPRs leading to loss of information due to a coverage probability higher than $1 - \alpha$ caused by the independence assumption.

One other correction that can be used is BH (Benjamin Hochberg) correction. It is typically used to control the false discovery rate. This would result in a prediction region where each period ($h$) in $H$ will have a different significance level, i.e. $\alpha_h = \alpha \times h/H$. The resulting prediction region would not be as conservative as the one obtained after Bonferroni's correction. However, they still assume independence and, therefore, are not ideal.

There are several other correction methods, including Sidak Correction, Holm's Step-Down procedure. All these methods can be used to string together marginal prediction intervals to construct a joint prediction region.

## 3.3 Wolf and Wunderli Joint Prediction Region

This section discusses Wolf and Wunderli JPR in detail and introduces k-FWE. The discussed JPR method is also referred to as k-FEW JPR.

### 3.3.1 FWE and k-FWE

Generally, the joint regions control the probability of containing the entire vector of interest to be at least equal to $1 - \alpha$. In other words, they control the probability of getting at least one component of the vector wrong to be at most equal to $\alpha$. This can be termed as familywise error rate ($FWE$).

$FWE = P\{at\ least\ one\ of\ the\ H\ components\ not\ contained\ in\ the\ joint\ region\}$

This proves to be a very stringent definition, and controlling for $FWE$ becomes strict with increasing H. Wolf and Wunderli paper suggests a generalised $FWE$, also called $K - FWE$, which provides flexibility in the error-rate criteria.

$k - FWE = P\{at\ least\ k\ of\ the\ H\ components\ not\ contained\ in\ the\ joint\ region\}$

Using $k - FWE$ lets the researcher decide the relaxation of the joint region according to the application, as now we are deeming some errors acceptable. $FWE$ can be seen as a special case of $k - FWE$ with $k = 1$. With $k > 1$, the error rate becomes less stringent. It is intuitive to see that larger $k$ causes smaller joint regions due to accepting few errors or ignoring them.

### 3.3.2 Notations and Definitions for Univariate time-series

- Time Series: $\{y_1 \ldots \ldots y_T\}$

- Probability Generating Mechanism: **P**
- Estimated Probability Generating Mechanism: $\hat{P}_T$
- Future path to predict: $Y_{T,H} = (Y_{T+1}\ldots\ldots Y_{T+H})'$
- Path forecast: $\hat{Y}_T(H) = (\hat{y}_T(1)\ldots\ldots\hat{y}_T(H))'$
- Vector of Prediction errors: $\hat{U}_T(H) = (\hat{u}_T(1)\ldots\ldots\hat{u}_T(H))' = \hat{Y}_T(H) - Y_{T,H}$
- Prediction standard error: $\hat{\sigma}_T(h)$ i.e., the standard error for $\hat{u}_T(h)$
- Bootstrapped prediction errors: $\hat{U}_T^*(H) = (\hat{u}_T^*(1)\ldots\ldots\hat{u}_T^*(H))'$
- Bootstrapped Data $= \{y_1^*\ldots\ldots y_T^*, y_{T+1}^*\ldots\ldots y_{T+H}^*\}$
- Bootstrap-Prediction standard error: $\hat{\sigma}_T^*(h)$ i.e., the standard error for $\hat{u}_T^*(h)$
- Standardised prediction errors: $\hat{S}_T(H) = (\hat{u}_T(1)/\hat{\sigma}_T(1)\ldots\ldots\hat{u}_T(H)/\hat{\sigma}_T(H))'$
- Bootstrapped Standardised prediction errors: $\hat{S}_T^*(H) = (\hat{u}_T^*(1)/\hat{\sigma}_T^*(1)\ldots\ldots\hat{u}_T^*(H)/\hat{\sigma}_T^*(H))'$
- Probability Law under **P** of $\hat{S}_T(H)|y_T, y_{T-1}\ldots = \hat{J}_T$
- Probability Law under $\hat{P}_T$ of $S_T^*(H)|y_T^*, y_{T-1}^*\ldots = \hat{J}_T^*$

The way the above equations are used is that first, we take the observed time-series $\{y_1\ldots\ldots y_T\}$ to create a model to make the predictions $\hat{y}_T(h)$ for every $h$ in $H$. This fitted model is used to find prediction standard error $\hat{\sigma}_T(h)$. Then, the model is used to generate bootstrapped data $\{y_1^*\ldots\ldots \hat{y}_T^*, y_{T+1}^*\ldots\ldots \hat{y}_{T+H}^*\}$. If we do not have confidence in our fitted model, we can also use non-parametric bootstrap like block bootstrap or sieve bootstrap. Prediction standard error $\hat{\sigma}_T^*(h)$ i.e. standard error of the predictions can be computed by not using the $\{y_{T+1}^*\ldots\ldots \hat{y}_{T+H}^*\}$ and finally, bootstrapped prediction errors can be computed.

### 3.3.3 Defining K-FWE JPR

- $k - max(X)$: If we suppose $X$ as a vector of length $X$, then $k - max(X)$ is the $K^{th}$ largest element of the vector $X$.
- If we organise all the components of the vector $X$ in increasing order, then the $k - max(X)$ equals $X_{(H-k+1)}$ element.
- $k - min(X)$: If we suppose $X$ is a vector of length $L$, then $k - min(X)$ is the $k^{th}$ smallest element of the vector $X$. If we organise all the components of the vector $X$ in increasing order, then the $k - min(X)$ equals $X_{(k)}$ element.
- $|X|$ denotes the absolute value of the vector/all components of the vector.
- $d_{|.|,1-\alpha}^{k-max} := 1 - \alpha$ $quantile\ of\ the\ random\ variable\ k - max(|\hat{S}_T(H)|)$

Using it, we can find the two-sided JPR for $Y_{T,H}$ that controls $k - FWE$ in finite samples, which is

$$[\hat{y}_T(1) \pm d_{|.|,1-\alpha}^{k-max} \cdot \hat{\sigma}_T(1)] \times \cdots\cdots \times [\hat{y}_T(H) \pm d_{|.|,1-\alpha}^{k-max} \cdot \hat{\sigma}_T(H)]$$

Theoretically, the probability that the above region will contain at least $H - k + 1$ elements of $Y_{T,H}$ is equal to at least $1 - \alpha$ in finite samples. This property can be inferred directly from the definition of the multiplier $d^{k-max}_{|.|,1-\alpha}$.

However, the above region is not feasible due to the unavailability of the multiplier $d^{k-max}_{|.|,1-\alpha}$ as the multiplier is not a functional. This multiplier can be estimated by $d^{k-max,*}_{|.|,1-\alpha}$, which is $1 - \alpha$ quantile of the random variable $k - max(|\hat{S}^*_T(H)|)$. This quantile cannot be derived analytically but can be simulated to arbitrary precision using bootstrapping with sufficiently large samples. Therefore, the two-sided JPR can be defined as:

$$[\hat{y}_T(1) \pm d^{k-max,*}_{|.|,1-\alpha} \cdot \hat{\sigma}_T(1)] \times \cdots \times [\hat{y}_T(H) \pm d^{k-max,*}_{|.|,1-\alpha} \cdot \hat{\sigma}_T(H)]$$

This implies that the region will contain at least $H - k + 1$ elements of $Y_{T,H}$ is equal to at least $1 - \alpha$ asymptotically.

The one-sided JPR's are also found similarly. With the one-sided lower JPR that controls $k - FWE$ asymptotically given by:

$$[\hat{y}_T(1) - d^{k-max,*}_{1-\alpha} \cdot \hat{\sigma}_T(1), \infty) \times \cdots \times [\hat{y}_T(H) - d^{k-max,*}_{1-\alpha} \cdot \hat{\sigma}_T(H), \infty)$$

Where $d^{k-max,*}_{1-\alpha}$ denotes $1 - \alpha$ quantile of the random variable $k - max(\hat{S}^*_T(H))$.

Similarly, one-sided upper JPR that controls $k - FWE$ asymptotically is given by:

$$(-\infty, \hat{y}_T(1) - d^{k-min,*}_{\alpha} \cdot \hat{\sigma}_T(1)] \times \cdots \times (-\infty, \hat{y}_T(H) - d^{k-min,*}_{\alpha} \cdot \hat{\sigma}_T(H)]$$

Where $d^{k-min,*}_{\alpha}$ denotes $\alpha$ quantile of the random variable $k - min(\hat{S}^*_T(H))$. It's worth noting that this multiplier is a negative number; therefore, the upper end of the interval is a larger number than the point prediction.

The multipliers $d^{k-max,*}_{1-\alpha}$ and $d^{k-max,*}_{|.|,1-\alpha}$ are monotonically decreasing in $K$, whereas $d^{k-min,*}_{\alpha}$ is monotonically increasing in $k$.

### 3.3.4 Algorithm for JPR computation for univariate time-series

1. Generate bootstrap data $\{y^*_1 \ldots y^*_T, y^*_{T+1} \ldots y^*_{T+H}\}$
2. Use T observations $\{y^*_1 \ldots y^*_T,\}$ to compute forecasts $\hat{y}^*_T(h)$ and prediction standard errors $\hat{\sigma}^*_T(h)$.
3. Compute prediction standard errors $u^*_T(h) = \hat{y}^*_T(h) - y^*_{T+h}$.
4. Compute standardised prediction errors $\hat{s}^*_T(h) = \hat{u}^*_T(h)/\hat{\sigma}^*_T(h)$ and $\hat{S}^*_T(H) = (\hat{u}^*_T(1)/\hat{\sigma}^*_T(1) \ldots \hat{u}^*_T(H)/\hat{\sigma}^*_T(H))'$
5. Compute $k - max^* = k - max(\hat{S}^*_T(H))$, $k - max^*_{|.|} = k - max(|\hat{S}_T(H)|)$, $k - min^*_{|.|} = k - min(\hat{S}^*_T(H))$.
6. Repeat steps 1 to 5, $B$ times, resulting in the statics in step 5 $B$ times.

7. Find the relative multiplier as $1 - \alpha$ or $\alpha$ quantile based on the band

### 3.3.5 Notations and Definitions for multivariate time-series

The above algorithm can be applied simply to a multivariate time series with some minor notational changes, other notations remain the same as the univariate case.

- Time Series: $\{Z_1 \ldots \ldots Z_T\}$ with $Z_t = (z_{1,t}, \ldots, z_{v,t})'$ where $v$ is the components of the time series.
- To avoid loss of generality, we wish to predict the first component of $Z_t$ and write $Z_t = (y_1, z_{2,t} \ldots, z_{v,t})'$
- Path forecast: $\hat{Y}_T(H) = (\hat{y}_T(1) \ldots \ldots \hat{y}_T(H))'$ where $\hat{Y}_T(H)$ is now a function of $Z_T$ now.
- Probability Law under $P$ of $\hat{S}_T(H)|Z_T, Z_{T-1}\ldots = \hat{J}_T$
- Probability Law under $\hat{P}_T$ of $S_T^*(H)|Z_T^*, Z_{T-1}^*\ldots = \hat{J}_T^*$

### 3.3.6 Algorithm for multivariate time-series

1. Generate bootstrap data $\{Z_1^* \ldots \ldots Z_T^*, Z_{T+1}^* \ldots \ldots Z_{T+H}^*\}$.
2. Use T observations $\{Z_1^* \ldots \ldots Z_T^*\}$ to compute forecasts $\hat{y}_T^*(h)$ and prediction standard errors $\hat{\sigma}_T^*(h)$.
3. Follow steps 4,5,6,7 from the univariate case above.

### 3.3.7 Features of k-FWE JPR

The k-FWE JPR has some salient features, making it more attractive than the others. Unlike Scheffé JPR and Jordà and Marcellino, it does not assume the normality of the vector of prediction error or convergence to mean zero.

It makes a high-level assumption to construct the JPR. It assumes that the conditional distribution of the vector of standardised bootstrapped prediction errors $\hat{S}_T^*(H)$ becomes more and more reliable approximation to the (unknown) conditional distribution of the vector of true standardised prediction errors $\hat{S}_T(H)$ as the Sample size T increases. It can be formally stated as $\hat{J}_T$ converges to a non-random continuous limit law $\hat{J}$. Furthermore, $\hat{J}_T^*$ consistently estimates this limit law: $\rho(\hat{J}_T^*, \hat{J}_T) \to 0$, for any metric $\rho$ implying weak convergence.

The assumption made ensures that all the constructed JPR's (two-sided, one-sided) for future paths satisfies

$$\limsup_{T \to \infty} \boldsymbol{k - FWE} \leq \alpha$$

The assumption ensures asymptotic validity, but it does not address the finite-sample property. For ensuring best-possible finite sample performance, it is needed to have the distribution of $\hat{J}_T^*$ is as close as possible to the distribution of $\hat{J}_T$. A bootstrapping method to address finite-sample performance is discussed later.

It provides the balanced property, whose desirability was discussed in Staszewska-Bystrova (2013) JPR section.

Under the additional assumption that the marginal distribution of

$$(\hat{y}_T(h) - y_{T,h})/\hat{\sigma}_T(h)$$

is the same for all $h = 1, \ldots, H$, asymptotically. Then, the probability

$$\boldsymbol{P}\{y_{T+h}\boldsymbol{\epsilon}[\hat{y}_T(h) \pm d_{|.|,1-\alpha}^{k-max,*} \cdot \hat{\sigma}_T(h)]\}$$

is the same for all $h = 1, \ldots, H$, asymptotically. This way, all the forecasts are treated equally.

This additional assumption holds if the time series is generated by ARIMA models with IID errors, for any reasonable model-based way to compute the forecasts $\hat{y}_T(h)$ and the prediction errors $\hat{\sigma}_T(h)$.

As discussed earlier, k-FWE JPR is calculated directly in a rectangular shape and is not formed by conversion from an elliptical JPR. Hence, it ensures no loss of information that occurred when an elliptical region is transformed into a rectangular one, leading to conservative JPR.

### 3.3.8 Computing prediction standard errors

Apart from deriving or computing the multiplier, which controls the volume of the region and k-FWE, the other important quantity that needs estimating for calculation of JPR is the prediction standard errors $\hat{\sigma}_T(h)$. Wolf et al. paper estimated the prediction standard error using the Box-Jenkins method. The derived prediction standard errors are only suitable for $AR(p)$ models, and applying other models would require finding different ways of estimation for the prediction standard errors.

One simple approach that can work regardless of any model is bootstrapping to find prediction standard errors. Since bootstrapping is already an integral part of the JPR computation, it would be pretty straightforward to use them and get the prediction standard errors $\hat{\sigma}_T(h)$. Utilising those bootstraps to get prediction standard errors does not require any more computation. However, it's not only prediction standard errors that need computing but also bootstrapped prediction errors $\hat{\sigma}_T^*(h)$ which is the prediction standard error of a bootstrapped sequence. The paper estimated them in the same way as prediction standard error. Just as bootstrapping is suggested to estimate the prediction standard error, it can be applied for the estimation of bootstrapped prediction errors $\hat{\sigma}_T^*(h)$. This method of using bootstrapping for a bootstrapped sequence is called double bootstrapping. However, it causes a minor problem as using double bootstrapping increases computational time exponentially. Theoretically, a bootstrapped series is a proxy for our sample. If our series length/ sample size is big enough and the applied bootstrapping method is sound, their statistics should be approximately the same. A shortcut would be to assume that the calculated bootstrapped prediction standard errors would be very close to prediction standard errors saves from the computation. However, the results obtained are not as accurate as those obtained after double bootstrapping, and the strength of assumption violation decides the accuracy of results.

## 3.4 Notations, Definitions, and Scheffé JPR and JCR

We are using Single Forecast to define notation and use it as a building block for later discussion as done in Wolf and Wunderli paper[1].

Let's consider a random variably **y** with mean $\boldsymbol{\mu} := \mathbf{E(y)}$ as a special case before moving towards a more general case of a random vector with H elements. We may wish to estimate $\boldsymbol{\mu}$ or to predict **y**.

$\hat{y}$ − forecast of **y**

$\hat{\mu}$ − estimator of $\boldsymbol{\mu}$

**P** − underlying probability mechanism

In most cases, both quantities are the same, i.e., $\hat{y} = \hat{\mu}$, for example, in the case of linear regression with quadratic loss. However, we look for 'uncertainty interval' it differs for both quantities. The former is a random variable, and the interval needs to be wider to accommodate that randomness, whereas the latter is not a random variable and has a narrower interval. For the purpose of this thesis to make the distinction clear, the prediction interval is used for the former quantity and the confidence interval for the latter.

The above notation can easily be extended to a random vector with each component as a variable with $\boldsymbol{Y} := (y_1 \ldots y_H)'$ and $\mu := (\mu_1 \ldots \mu_H)'$. For mathematical purposes, **Y** will correspond to the value of random variable 1 to H periods into the future. Just like the case for a single random variable, random vectors can be denoted by:

$\hat{Y}$ − forecast of **Y**

$\hat{\boldsymbol{\mu}}$ − estimator of $\boldsymbol{\mu}$

The two quantities are the same as earlier. The same logic of wider region for the former and narrower for the latter also applies due to additional randomness in the former.

A potential complication occurs when we desire uncertainty statements about individual components $y_h$ and $\mu_h$. For example, when the JPR for **Y** is required along with the path forecast $\hat{Y}$. The aim is to find the upper and the lower confidence bounds for every $y_h$ such that the probability of containing the true sequence is $1 - \alpha$. This is a trivial problem if the resulting region is rectangular, but it is not always possible; many methods in the past have had regions in spherical or elliptical form. One such important example is the Scheffé joint region, which used the inversion of the classical F-test[15] to find the desired region.

Scheffé joint region

$$JCR := \{\boldsymbol{\mu}_o : (\hat{\boldsymbol{\mu}} - \boldsymbol{\mu}_o)' [\hat{\boldsymbol{\Sigma}}(\hat{\boldsymbol{\mu}})] (\hat{\boldsymbol{\mu}} - \boldsymbol{\mu}_o) \leq \chi^2_{H, 1-\alpha}\}$$

$$JPR := \left\{X: (\widehat{Y} - X)' [\widehat{\Sigma}(\widehat{U})](\widehat{Y} - X) \leq \chi^2_{H,1-\alpha}\right\}$$

The above formulas define confidence region and prediction regions, respectively, where $\chi^2_{H,1-\alpha}$ is the $1 - \alpha$ quantile of the chi distribution with $H$ degrees of freedom.

The use of JCR is justified by using the central limit theorem implying normal distribution of the parameter $\widehat{\boldsymbol{\mu}}$ with mean $\boldsymbol{\mu}$. The calculation of JCR is reasonable as the distribution of parameters is often normal. This will hold under mild regularity condition. The JPR is defined similarly, with $\widehat{U}$ as a vector of prediction errors $\widehat{Y} - Y$ and $\widehat{\Sigma}(\widehat{U})$ as the covariance matrix of the prediction errors. The use of this JPR is only justified if the vector of prediction errors is normally distributed with mean 0, which is an overly strong assumption and is often violated in practice. While the central limit theorem can be applied to show that an estimator has an approximately normal distribution, it fails to show that the forecast error has a normal distribution for large sample sizes.

If the joint region is of elliptical/spherical form and a statement concerning individual components is desired, the joint region must be projected onto the axes $\boldsymbol{R^H}$. However, doing this creates a larger rectangular joint region. It will be the smallest rectangle that can inscribe the spherical/elliptical region and would result in a region with a joint coverage probability greater than $1 - \alpha$. This bound will be a very conservative bound. Suppose we want a statement concerning an individual component with a joint coverage probability $1 - \alpha$. In that case, it makes sense to create a 'direct' rectangular joint region instead of converting an elliptical/spherical region to a rectangular one.

### 3.4.1 Jordà and Marcellino(2010) JPR

Jordà and Marcellino (2010)[8] proposed an asymptotic method to create a Joint prediction region(JPR). The JPR is given by:

$$\widehat{y}_T(H) \pm P \left[\sqrt{\chi^2_{h,1-\alpha}/h}\right]_{h=1}^{H}$$

Where P is the lower triangular Cholesky decomposition of $\widehat{\Xi}_H/T$, ($\widehat{\Xi}_H$ is the estimated covariance matrix ).

$\left[\sqrt{\chi^2_{h,1-\alpha}/h}\right]_{h=1}^{H}$ is an $H \times 1$ vector, where $\chi^2_{h,1-\alpha}$ denotes the $1 - \alpha$ quantile of the chi-square distribution with $h$ degrees of freedom.

**Definition 3.3 (Cholesky Decomposition)** Cholesky Decomposition is a method to decompose a Hermitian, positive-definite matrix into the product of two matrices i.e. lower triangular matrix and its transposed conjugate matrix. It is written as:

$$H = P \times P'$$

where $H$ is a Hermitian, positive-definite matrix.

The defined JPR is based on the assumption that

$$\sqrt{T}\bigl(\hat{Y}_T(H) - Y_{T,H} | Z_T, Z_{T-1}\bigr) \xrightarrow{d} N(\mathbf{0}, \Xi_H)$$

Where $\xrightarrow{d}$ denotes convergence in distribution and on the availability of a consistent estimator $\hat{\Xi}_H \xrightarrow{P} \Xi_H$, where $\xrightarrow{P}$ denotes convergence in probability.

**Problems with this JPR**

The assumption is strong and problematic for various reasons. It implies that the conditional distribution of the vector of prediction error $\hat{U}_T(H) := \hat{Y}_T(H) - Y_{T,H}$ is a multivariate Gaussian with a mean of zero, at least with large values of T. This is unrealistic as the conditional distribution of prediction error is highly dependent on the conditional distribution of the random variable to be predicted, which is generally not gaussian. It also implies that with large values of T, the conditional distribution of prediction errors converges weakly to a point mass of zero. This implication could be true for the difference between the population parameter and the estimator, but it is unrealistic for the vector of prediction errors. Even if all the model parameters are known, forecasting can not be done accurately due to its random nature. A simple example of $AR(1)$ model can easily show it.

Let's consider the true model as $y_t = \rho y_{t-1} + \epsilon_t$

Where $|\rho| < 1$ and the errors $\{\epsilon_t\}$ are IID with mean zero and finite variance $\sigma_\epsilon^2 > 0$. At time $T$, the forecast of $y_{T+1}$ is

$$\hat{y}_{T+1} := \hat{v} + \hat{\rho} y_T$$

Where $\hat{v}$ and $\hat{\rho}$ are consistent estimators of $v$ and $\rho$ respectively.

$$\hat{u}_T(1) = \hat{v} + \hat{\rho} y_T - y_{T+1}$$

As $T$ approaches infinity, the conditional distribution of $\hat{u}_T(1)$ converges to the distribution of $-\epsilon_{T+1}$ and is not dependent on $T$. The distribution is neither gaussian nor a point mass at zero.

One other problem with this JPR is that its width is not necessarily increasing with forecast horizon $h$ because the multiplier, $\sqrt{\chi^2_{h,1-\alpha}/h}$, is monotonically decreasing in $h$.

Another problem with this JPR was shown by Staszewska-Bystrova (2013)[10], suggesting that if the matrix $P$ presents negative values, it must be replaced by $|P|$ (absolute values of matrix $P$). This makes the defined region as

$$\hat{y}_T(H) \pm |P| \left[\sqrt{\chi^2_{h,1-\alpha}/h}\right]_{h=1}^{H}$$

The way the JPR is derived is heuristic and lacks theoretical justification. This JPR is derived from Scheffé JPR and, therefore, is referred to as Modified Scheffé JPR.

### 3.4.2 Staszewska-Bystrova (2011) JPR

Staszewska-Bystrova (2011)[9] proposed an alternative bootstrap method to control FWE (Family wise error). The method is purely heuristic and does not provide any asymptotic validity under some high-level assumption. Due to its heuristic nature, the JPR is referred to as NP heuristic JPR.

First, conditional on the observed data, B bootstrap path forecasts are generated, i.e $\hat{Y}_T^{*,b}(H)$, for $B = 1, \dots, B$. Then, $\alpha B$ of these bootstrapped forecasts are discarded: namely $\hat{Y}_T^{*,b}(H)$, which are furthest away from the real path $\hat{Y}_T(H)$, where the distance between the $H \times 1$ vectors is calculated using Euclidean distance. The distance metric was chosen experimentally. This creates the final JPR as the envelope of remaining $(1 - \alpha)B$ path forecasts. The method to use neighbouring paths for constructing JPR worked well, but it has some concerns apart from the lack of asymptotic validity.

The proposed method is restricted to (V)AR models as it used the backward representation of (V)AR model to generate bootstrap path forecasts. One problem that backward representation causes is that when the forward errors are non-gaussian, even if they are independent, the backward errors are not independent but merely uncorrelated. Therefore, using backward representation to create bootstrap on the residuals may not always be valid.

One problem in the proposed method is the use of Efron's percentile method[17] equivalents to 'looking up the wrong tail of a distribution.' Theoretically, it is argued that such a method can only work if the conditional distribution of forecast error is normally distributed and is centred around zero. The method's performance may suffer when the vector of forecast error is not multivariate normal and is skewed.

The JPR constructed is not balanced. In other words, the JPR obtained after discarding $\alpha B$ forecasts has a jagged shape and can be considered unattractive. A JPR is considered balanced if the length of all the prediction intervals is the same for all the observations in $H$.

One way to show why the balanced property is desirable is by considering the JPR below:

$$PI_T(1) \times (-\infty, \infty) \times \dots \times (-\infty, \infty),$$

Where $PI_T(1)$ is the prediction interval for $y_{T+1}$ with significance level α. Despite being a bad JPR, it has the property of containing all the future paths with a coverage probability of $1 - \alpha$, asymptotically. It only requires the first prediction interval to satisfy the coverage probability constraint. This consolidates the claim that the property of balance is desirable and attractive as it can consider all the forecasts in the future path equally important.

# 4 Results

This section highlights the results obtained after comparing different predictors and different JPR methods on different datasets. Different datasets have different properties, and the results vary a little. Reiterating that one of the assumptions made is that the prediction standard error, $\hat{\sigma}_T(h)$, and the bootstrapped prediction errors, $\hat{\sigma}_T^*(h)$, are approximately equal. This violation of the assumption depends on the chosen dataset and the number of bootstrap samples along with the chosen bootstrapping method. The assumption saves a lot of computation time. To provide an example, the LSTM model for the dataset in section 6.1 took almost 6 hours to provide empirical coverage with $10^4$ bootstrapped predictions, the proposed method to use double bootstrapping may take 600 hours if 100 bootstrapped samples are used for every single bootstrapped sample.

## 4.1 Synthetic dataset

This dataset is a synthetic dataset, and the code to produce it is provided separately. The code is adapted from one of the exercises on Coursera's TensorFlow Developer course I studied[5]. The Synthetic data has 3651 observations with seasonality of 30. Using synthetic data offers the advantage of finding a model quickly. It also follows a certain pattern and is easier to bootstrap. The figure below shows a plot of the data with the first 120 observations.

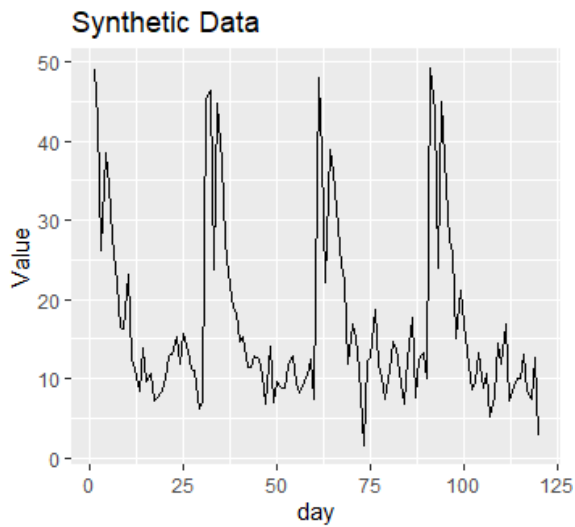

**Figure 5:** Plot for synthetic data

This data is used to find empirical coverages and empirical widths for the predictions with different JPR methods and Predictors. For a given setting, a JPR is constructed using approximately 2225 observations for the next $H$ predictions. The outcome is judged as a success if all but at most $k-1$ of $H$ observations are contained in the JPR. The JPR's are constructed 100 times, and the number of successes gives empirical coverage. For example, if $k = 1$, and emp. coverage is 100 it means that all the 100 prediction sequences are included in the derived JPR; if $k =$ 2 and emp. coverage is 100, which means at least $H - k + 1$ observations of every prediction sequence are included in the derived JPR. The data is used in a rolling window fashion. The first 2245 observations are used to predict next $H$ observations for the first simulation; the second simulation takes a lag of 10 from the previous simulation and uses the last 2245 observations to predict the next $H$ observations and so on.

The empirical width is calculated similarly; the geometric width for each JPR in the rolling window is calculated and is averaged over all rolling windows to provide a good estimate for the dataset.

The geometric-average of widths for a sequence JPR of given $H$ can be written as

$$w_{geom} \coloneqq (\Pi_{h=1}^{H} w_h)^{(1/H)}$$

Where $w_h$ is the width for a particular $H$. The Wolf et al. paper uses the same methodology to provide a fair assessment of the method's out-of-sample performance. The paper also mentioned that the method is not very accurate because the series are not independent.

One important point to note is that the coverages and results obtained are only after using 1000 bootstrapped samples and 100 runs; this makes it hard to make accurate empirical averages. For further elaboration, an unbiased coin is supposed to have 50 heads and 50 tails for every 100 tosses, but it seldom happens the divide might be 46-54 or 55-45. The true average is found when many of these experiments are conducted and averaged. For the JPR, it might be done using Monte-Carlo simulations, which provide a better estimate than the estimate from practical application. However, practical applications don't always provide the same results like the ones obtained from Monte-Carlo simulations.

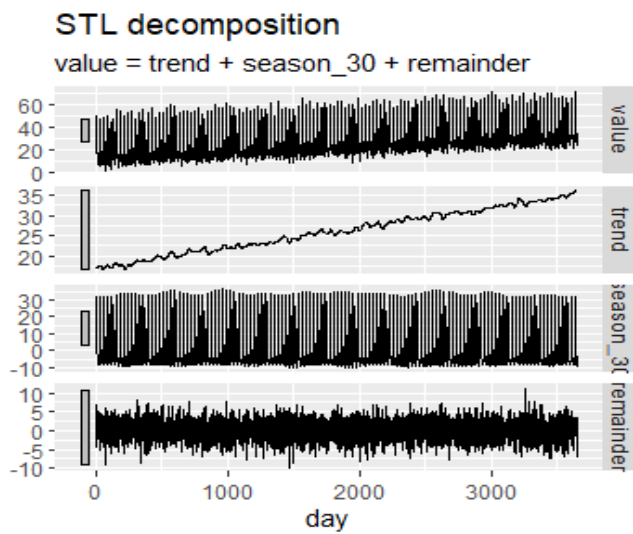

**Figure 6** STL decomposition of Synthetic Data

The applied bootstrapping technique uses STL decomposition to decompose the series into relevant components, and R's generate function is used to create a bootstrap sample. The figure below shows STL decomposition for the given dataset, consisting of a trend, a seasonal, and an error term.

### 4.1.1 ETS model

The first results are for ETS (Exponential Smoothing) model. The model used does not account for seasonality and is usually works better for one-step predictions. The result is that the resulting forecast has a high error rate, leading to very large prediction intervals. Results are mainly compared for $\alpha = (0.1, 0.2, 0.3)$, $k = (1,2,3)$, $H = (6,12,18,24)$ for k-FWE JPR against Joint Marginals with Bonferroni's correction. The table below shows the corrected coverages with different values of $H$.

| BF Correction | $H = 6$ | $H = 12$ | $H = 18$ | $H = 24$ |
|---|---|---|---|---|
| $1 - \alpha = 0.9$ | 0.983 | 0.991 | 0.994 | 0.995 |
| $1 - \alpha = 0.8$ | 0.966 | 0.983 | 0.988 | 0.991 |
| $1 - \alpha = 0.7$ | 0.95 | 0.975 | 0.980 | 0.987 |

**Table 1**: Equivalent 1-α after Bonferroni Correction

As the above table summarises, Bonferroni correction is very conservative, and a desired coverage of 0.7 using it for JPR computation requires marginal coverage to be at least 0.95 when $H$ is greater or equal to 6. Theoretically, we can expect Joint Marginals to provide more coverage than desired due to no independence between observations.

| Emp. Coverage, $1 - \alpha = 90\%$ | $H = 6$ | $H = 12$ | $H = 18$ | $H = 24$ |
|---|---|---|---|---|
| Joint marginals (Bonferroni's correction) | 89 | 93 | 97 | 100 |
| 1-FWE JPR | 87 | 86 | 86 | 86 |
| 2-FWE JPR | 93 | 89 | 88 | 88 |
| 3- FWE JPR | 91 | 88 | 88 | 87 |

**Table 2**: Empirical coverage for $1 - \alpha = 90\%$

| Emp. Coverage, $1 - \alpha = 80\%$ | $H = 6$ | $H = 12$ | $H = 18$ | $H = 24$ |
|---|---|---|---|---|
| Joint marginals (Bonferroni's correction) | 79 | 87 | 94 | 96 |
| 1-FWE JPR | 78 | 73 | 76 | 79 |
| 2-FWE JPR | 77 | 78 | 75 | 74 |
| 3- FWE JPR | 79 | 78 | 76 | 78 |

**Table 3**: Empirical coverage for $1 - \alpha = 80\%$

| Emp. Coverage, $1-\alpha = 70\%$ | $H = 6$ | $H = 12$ | $H = 18$ | $H = 24$ |
|---|---|---|---|---|
| Joint marginals (Bonferroni's correction) | 73 | 84 | 90 | 93 |
| 1-FWE JPR | 68 | 65 | 67 | 64 |
| 2-FWE JPR | 65 | 69 | 66 | 67 |
| 3- FWE JPR | 68 | 68 | 66 | 66 |

**Table 4**: Empirical coverage for $1-\alpha = 70\%$

| Emp. Width, $1-\alpha = 90\%$ | $H = 6$ | $H = 12$ | $H = 18$ | $H = 24$ |
|---|---|---|---|---|
| Joint marginals (Bonferroni's correction) | 68.45776 | 85.63997 | 101.17732 | 112.30706 |
| 1-FWE JPR | 53.17210 | 57.03878 | 79.23678 | 79.80063 |
| 2-FWE JPR | 25.22664 | 49.23422 | 67.28866 | 68.77467 |
| 3- FWE JPR | 21.32554 | 45.54914 | 51.50198 | 63.98194 |

**Table 5**: Empirical Width for $1-\alpha = 90\%$

| Emp. Width, $1-\alpha = 80\%$ | $H = 6$ | $H = 12$ | $H = 18$ | $H = 24$ |
|---|---|---|---|---|
| Joint marginals (Bonferroni's correction) | 61.34609 | 78.18901 | 92.48588 | 104.76667 |
| 1-FWE JPR | 50.84540 | 54.79271 | 76.79351 | 77.43188 |
| 2-FWE JPR | 23.26170 | 47.29209 | 65.06058 | 66.68677 |
| 3- FWE JPR | 19.61782 | 43.58893 | 49.23127 | 61.89519 |

**Table 6**: Empirical Width for $1-\alpha = 80\%$

| Emp. Width, $1-\alpha = 70\%$ | $H = 6$ | $H = 12$ | $H = 18$ | $H = 24$ |
|---|---|---|---|---|
| Joint marginals (Bonferroni's correction) | 56.04771 | 73.45202 | 85.46863 | 99.59928 |
| 1-FWE JPR | 49.05882 | 53.16504 | 75.04036 | 75.65285 |
| 2-FWE JPR | 21.91274 | 45.91110 | 63.51061 | 65.14295 |
| 3- FWE JPR | 18.38535 | 42.28655 | 47.74107 | 60.45444 |

**Table 7**: Empirical Width for $1-\alpha = 70\%$

One important thing to note is that the Simple exponential Smoothing is chosen here, which means the forecast value is the same for all $h$, causing huge widths.

The results are the same as expected- Joint Marginals with BF correction provides the largest widths. It also has higher coverage than desired as the observations in the sequence is not independent of each other. One additional thing to notice is that the prediction intervals and, in turn, empirical width for BF correction are computed using bootstrapping parameters in the generate function with a value of 3000 bootstraps. The results vary if prediction intervals are calculated using the default method in the library, which provides symmetric intervals.

### 4.1.2 ARIMA model

The second model used is ARIMA or Seasonal ARIMA as the variant takes seasonality into account. Because they have seasonality component parameters, we expect them to work significantly better than the ETS model and provide lesser width or empirical width. The chosen ARIMA model has the form $pdq(0,0,1) + PDQ(0,1,0,30)$ where $pdq$ defines $AR(p)$, $Differencing(d)$, and $MA(q)$ for non-seasonal component and $PDQ$ defines $AR(P)$, $Differencing(D)$, $MA(Q)$, and seasonality ($S$) for the seasonal component.

| Emp. Coverage, $1-\alpha = 90\%$ | $H = 6$ | $H = 12$ | $H = 18$ | $H = 24$ |
|---|---|---|---|---|
| Joint marginals (Bonferroni's correction) | 92 | 94 | 91 | 93 |
| 1-FWE JPR | 93 | 88 | 91 | 88 |
| 2-FWE JPR | 86 | 84 | 88 | 85 |
| 3- FWE JPR | 88 | 82 | 83 | 83 |

**Table 8**: Empirical coverage for $1-\alpha = 90\%$

| Emp. Coverage, $1-\alpha = 80\%$ | $H = 6$ | $H = 12$ | $H = 18$ | $H = 24$ |
|---|---|---|---|---|
| Joint marginals (Bonferroni's correction) | 83 | 88 | 83 | 85 |
| 1-FWE JPR | 84 | 81 | 81 | 79 |
| 2-FWE JPR | 76 | 71 | 76 | 74 |
| 3- FWE JPR | 79 | 70 | 65 | 69 |

**Table 9**: Empirical coverage for $1-\alpha = 80\%$

| Emp. Coverage, $1-\alpha = 70\%$ | $H = 6$ | $H = 12$ | $H = 18$ | $H = 24$ |
|---|---|---|---|---|
| Joint marginals (Bonferroni's correction) | 78 | 84 | 75 | 79 |
| 1-FWE JPR | 67 | 72 | 76 | 72 |
| 2-FWE JPR | 70 | 60 | 60 | 66 |
| 3- FWE JPR | 72 | 63 | 58 | 55 |

**Table 10**: Empirical coverage for $1-\alpha = 70\%$

| Emp. Width, $1-\alpha = 90\%$ | $H = 6$ | $H = 12$ | $H = 18$ | $H = 24$ |
|---|---|---|---|---|
| Joint marginals (Bonferroni's correction) | 20.11879 | 22.26517 | 23.52642 | 24.10310 |
| 1-FWE JPR | 19.10870 | 20.97582 | 22.21315 | 23.00381 |
| 2-FWE JPR | 13.40294 | 16.01323 | 17.49112 | 18.41644 |
| 3- FWE JPR | 10.33879 | 13.31148 | 14.94680 | 16.01050 |

**Table 11**: Empirical Width for $1-\alpha = 90\%$

| Emp. Width, $1 - \alpha = 80\%$ | $H = 6$ | $H = 12$ | $H = 18$ | $H = 24$ |
|---|---|---|---|---|
| Joint marginals (Bonferroni's correction) | 17.85699 | 20.19292 | 21.30203 | 22.22888 |
| 1-FWE JPR | 16.76400 | 18.91553 | 20.22614 | 21.11980 |
| 2-FWE JPR | 11.80245 | 14.49201 | 16.04759 | 17.03850 |
| 3- FWE JPR | 8.931809 | 12.118276 | 13.794733 | 14.871342 |

**Table 12**: Empirical Width for $1 - \alpha = 80\%$

| Emp. Width, $1 - \alpha = 70\%$ | $H = 6$ | $H = 12$ | $H = 18$ | $H = 24$ |
|---|---|---|---|---|
| Joint marginals (Bonferroni's correction) | 16.52091 | 18.92757 | 19.63531 | 21.03719 |
| 1-FWE JPR | 15.17757 | 17.43733 | 18.85494 | 19.78939 |
| 2-FWE JPR | 10.71466 | 13.50525 | 15.01610 | 16.11177 |
| 3- FWE JPR | 7.938762 | 11.251646 | 12.959100 | 14.123382 |

**Table 13**: Empirical Width for $1 - \alpha = 70\%$

The results obtained following the same trend of higher empirical coverage and wider interval width with correction. One thing to note here is the huge difference between the predictor performance due to the seasonal component in the ARIMA model. It is also possible to use a variant of ETS (exponential smoothing) to account for the seasonality. However, its use is avoided here to show the difference in widths between poorly specified and well-defined models.

Another interesting aspect found after using ARIMA models is that the empirical coverages are falling short of desirable coverages with increasing $h$. We wish to reiterate the assumptions made at the beginning of the section and mention the recommended bootstrap samples are at least a thousand. The results using a hundred bootstrap samples are still a good indicator of the performance of the model.

**Note about NP heuristic**: For the Arima model, the NP heuristic method of Staszewska-Bystrova (2011)[9] is also applied. However, the coverage is not computed due to computation problems and fair comparison. The JPR method and NP heuristic method both use bootstrapping to compute prediction regions, but unlike JPR, the NP heuristic depends solely on bootstrapping to compute the regions. The JPR method has been shown to work well on the synthetic dataset, even with a bootstrapping of 100 samples. Still, the NP heuristic method is of a heuristic nature and is more sensitive to the number of bootstrapped samples requiring more bootstrapping than others. The NP heuristic did not provide good coverage properties during the initial experimentation as the width was very small. The number of bootstrap samples was increased to see the effect on width. As the width increased with more bootstrap samples, the few sequences became part of the regions. But with 10,000 samples, the same computation dilemma comes back. Considering, the JPR only used 100 bootstrap samples for prediction region calculation; it is unfair to draw comparisons if 10,000 samples are used for the NP heuristic. Still, it can be inferred that the NP heuristic method provides tight intervals, which follows from the theory itself. The NP heuristic is not considered in the coming sections for two reasons: Higher bootstrapping is needed than the JPR method, and coverage properties are not proved in the dissertation to draw comparisons. However, empirical coverages and empirical widths are calculated using 100 bootstrapped samples. The results showed less coverage than desired and smaller empirical widths. The empirical coverage and empirical width results along a width comparison with an increase in bootstrap size are provided in [Appendix section 9.1](#).

**Note about Modified Scheffé method**: Modified Scheffé JPR was also computed for the ARIMA model. The estimated covariance matrix was computed using 100 bootstrapped samples, and empirical coverages and widths were also obtained. The obtained empirical coverages were too small, and so were the widths. The main culprit seems to be the number of bootstrapped samples used for estimating the covariance matrix. It should be noted that it is recommended to at least use a thousand bootstrapped samples, and the results obtained in Wolf and Wunderli's paper[1] uses 10,000 bootstrapped samples whenever bootstrapping was required. However, even with 100 bootstrapped samples and assumptions, k-FWE JPR has shown promising results. Despite a small number of bootstrap samples, the expected difference between the JPRs is observed, with Modified Scheffé JPR performing the worst, followed by NP heuristic. These results are a good indicator of k-FWE JPR's reliability over Modified Scheffé JPR or NP heuristic JPR. The empirical coverages and empirical widths obtained after applying modified Scheffé JPR are provided in [Appendix section 9.2](#).

### 4.1.3 LSTM model

The third and final model used with the data set is the LSTM model. LSTM model is the first true machine learning model used in this thesis, others being considered statistical models. LSTM being an RNN, provides powerful predictions and require more data and computation. Ideally, it is desired to train all the simulation sequences separately and to get the respective predictions. However, this is highly unrealistic for Neural Networks, as they require more data and take a lot of time to train. Therefore, the whole synthetic data is used first to train the model, and the learnt parameters are used to get predictions for the rolling window dataset and all the bootstrapped sequences. The LSTM model is expected to outperform the models used before and is likely to provide narrower widths.

| Emp Coverage, $1-\alpha = 90\%$ | $H = 6$ | $H = 12$ | $H = 18$ | $H = 24$ |
|---|---|---|---|---|
| 1-FWE JPR | 92 | 91 | 87 | 88 |
| 2-FWE JPR | 90 | 90 | 94 | 93 |
| 3- FWE JPR | 87 | 84 | 91 | 89 |

**Table 14:** Empirical coverage for $1-\alpha = 90\%$

| Emp Coverage, $1-\alpha = 80\%$ | $H = 6$ | $H = 12$ | $H = 18$ | $H = 24$ |
|---|---|---|---|---|
| 1-FWE JPR | 80 | 82 | 75 | 76 |
| 2-FWE JPR | 74 | 74 | 82 | 75 |
| 3- FWE JPR | 74 | 75 | 77 | 80 |

**Table 15:** Empirical coverage for $1-\alpha = 80\%$

| Emp Coverage, $1-\alpha = 70\%$ | $H = 6$ | $H = 12$ | $H = 18$ | $H = 24$ |
|---|---|---|---|---|
| 1-FWE JPR | 70 | 69 | 67 | 66 |
| 2-FWE JPR | 66 | 62 | 69 | 66 |
| 3- FWE JPR | 69 | 65 | 62 | 71 |

**Table 16**: Empirical coverage for $1-\alpha = 70\%$

| Emp Width, $1-\alpha = 90\%$ | $H = 6$ | $H = 12$ | $H = 18$ | $H = 24$ |
|---|---|---|---|---|
| 1-FWE JPR | 16.54339 | 19.84922 | 21.58868 | 22.96380 |
| 2-FWE JPR | 10.90063 | 14.13094 | 15.98383 | 17.51132 |
| 3- FWE JPR | 8.031555 | 11.236610 | 13.198836 | 14.623274 |

**Table 17**: Empirical Width for $1-\alpha = 90\%$

| Emp Width, $1-\alpha = 80\%$ | $H = 6$ | $H = 12$ | $H = 18$ | $H = 24$ |
|---|---|---|---|---|
| 1-FWE JPR | 14.31257 | 17.52712 | 19.36685 | 20.76674 |
| 2-FWE JPR | 9.512208 | 12.654239 | 14.529825 | 15.991865 |
| 3- FWE JPR | 6.931526 | 10.101626 | 12.118382 | 13.533785 |

**Table 18**: Empirical Width for $1-\alpha = 80\%$

| Emp Width, $1-\alpha = 70\%$ | $H = 6$ | $H = 12$ | $H = 18$ | $H = 24$ |
|---|---|---|---|---|
| 1-FWE JPR | 12.81569 | 15.95303 | 17.87999 | 19.23954 |
| 2-FWE JPR | 8.594702 | 11.655545 | 13.583069 | 14.987987 |
| 3- FWE JPR | 6.172010 | 9.336591 | 11.329030 | 12.741107 |

**Table 19**: Empirical Width for $1-\alpha = 70\%$

As expected, LSTM has found better dependencies and has provided tighter width with good coverage properties.

**Note about Joint Marginals for LSTM:** Methods like ARIMA and ETS have inbuilt methods to find marginal prediction intervals. However, this is not automatically available for neural networks. One of the easiest ways is to use bootstrapping to find marginal prediction intervals. The method was used to find the JPR by using Joint Marginals with Bonferroni Correction. However, due to only 100 bootstrap samples, the estimation is crude and unreliable as this has caused under coverage whereas over coverage was expected. The results are provided in Appendix section 9.3.

The dataset has provided satisfactory results for the performance of JPR. The derived coverages are close to the desired level but not pinpoint, but this is expected due to assumptions made earlier in the chapter and relatively small sequences averaged over. A better estimate of the performance could be achieved using Monte-Carlo simulations, as also done in the paper, but that does not show the success in the real world.

## 4.2 Min. Temperature Dataset

I worked with this dataset for a weekly assignment during the Sequences, Time Series and Prediction course. The dataset contains daily minimum temperatures in Melbourne, Australia, recorded from 1981 to 1990. It has a total of 3652 observations. It is taken from the GitHub of Jason Brownlee [16], the author of MachineLearningMastery.com website. The dataset is available for public use. The dataset has clearly a yearly seasonality. It can be shown in the plot below.

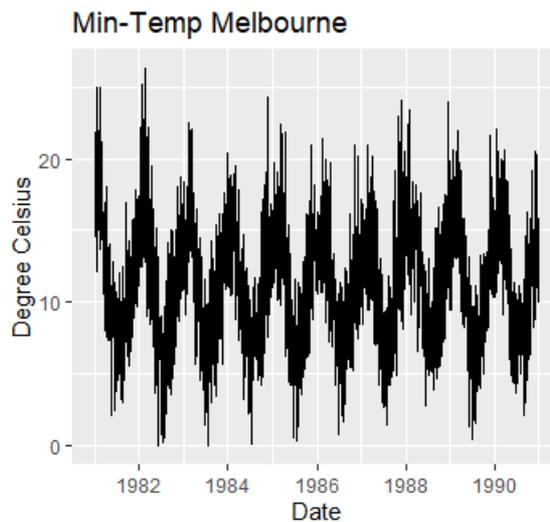

**Figure 7**: Plot of Min-Temp Dataset

The dataset has recorded min-temperatures for all but two days. Since the functions in fable libraries in R usually do not accept missing values, the values for the missing two observations are added by averaging the immediately previous and the next observation. Since they are only two observations, it does not affect the modelling except ensure that the functions in the fable package can be used. Let's look at the time-series decomposition for the dataset. The STL function has found two seasonality components in the dataset, one is yearly, which makes sense, but the other is weekly, which seems counter-intuitive. However, it looks plausible after looking at the graph for ten weeks, showing a cyclic pattern. After looking at bootstrapped sequences and the difference they made, we persisted with the decomposition provided by the STL function.

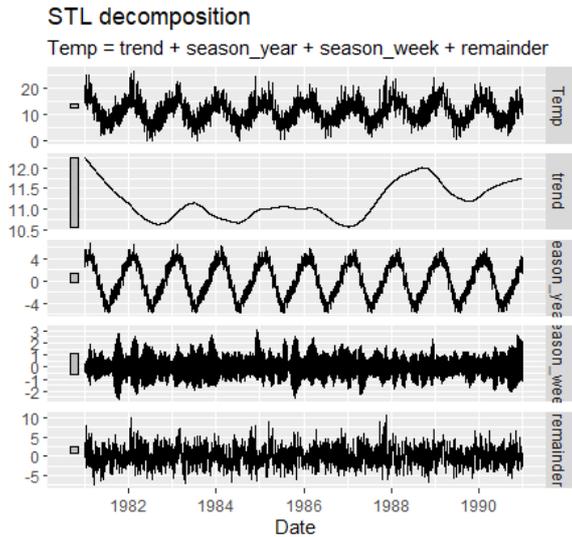 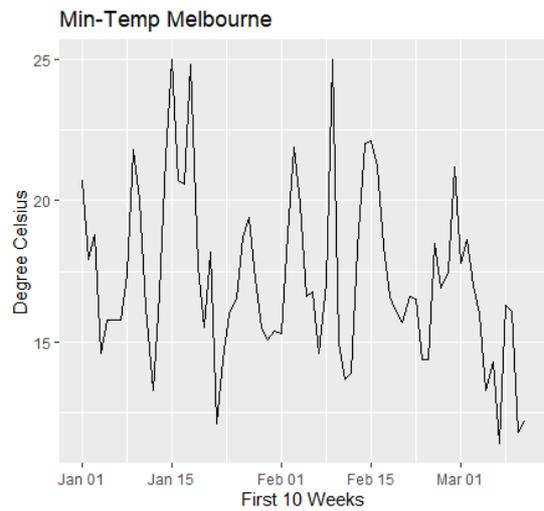

**Figure 8(a)** STL decomposition of Min-Temp     **Figure 8(b)** Plot for First 10 weeks

The prediction task is to predict the sequence of the next $H$ min-temp and the corresponding region. In the previous section, it is already proved that the k-FWE JPR has better convergence properties. This section has computed results only k-FWE JPR. Also, ETS has shown ineffectiveness without seasonality. The computed JPR's are only for Seasonal-ARIMA and LSTM based Neural Net. The calculated width calculated is only for the JPR of a given sequence and

### 4.2.1 ARIMA model

Just like in the previous section, the Seasonal variant of ARIMA is used. The chosen ARIMA model has the form $pdq(5,0,0) + PDQ(1,0,0,7)$ where $pdq$ defines $AR(p)$, $Differencing(d)$, and $MA(q)$ for the non-seasonal component and $PDQ$ defines $AR(P)$, $Differencing(D)$, $MA(Q)$, and seasonality ($S$) for the seasonal component. The ARIMA model is not defined this time and is chosen automatically.

| Emp Width, $1 - \alpha = 90\%$ | $H = 6$ | $H = 12$ | $H = 18$ | $H = 24$ |
|---|---|---|---|---|
| 1-FWE JPR | 13.67991 | 18.96266 | 19.51300 | 25.83318 |
| 2-FWE JPR | 10.30112 | 13.48160 | 15.24351 | 20.60463 |
| 3-FWE JPR | 7.258796 | 10.598606 | 12.293634 | 16.580643 |

**Table 20**: Prediction Width for $1 - \alpha = 90\%$

| Width, $1 - \alpha = 80\%$ | $H = 6$ | $H = 12$ | $H = 18$ | $H = 24$ |
|---|---|---|---|---|
| 1-FWE JPR | 11.83094 | 16.23604 | 16.80301 | 22.84962 |
| 2-FWE JPR | 8.569889 | 11.577976 | 13.178156 | 17.608404 |
| 3-FWE JPR | 6.026590 | 9.288439 | 11.037535 | 14.565222 |

**Table 21**: Prediction Width for $1 - \alpha = 80\%$

| Width, $1 - \alpha = 70\%$ | $H = 6$ | $H = 12$ | $H = 18$ | $H = 24$ |
|---|---|---|---|---|
| 1-FWE JPR | 10.60517 | 14.54753 | 15.32380 | 20.58172 |
| 2-FWE JPR | 7.423317 | 10.722313 | 11.967116 | 15.945728 |
| 3-FWE JPR | 5.121220 | 8.430370 | 9.922435 | 13.281978 |

**Table 22**: Prediction Width for $1 - \alpha = 70\%$

| | ARIMA | | | LSTM | | | |
|---|---|---|---|---|---|---|---|
| $h$ | lower | upper | prediction | lower | upper | prediction | True value |
| 1 | 9.15 | 26.62 | 17.88 | 7.19 | 27.40 | 17.29 | 15.5 |
| 2 | 10.27 | 24.25 | 17.26 | 8.99 | 23.76 | 16.38 | 14.1 |
| 3 | 10.62 | 24.04 | 17.33 | 9.33 | 22.66 | 15.99 | 11.0 |
| 4 | 10.57 | 23.76 | 17.17 | 9.44 | 22.04 | 15.74 | 11.1 |
| 5 | 10.44 | 23..33 | 16.88 | 9.79 | 21.44 | 15.62 | 14.0 |
| 6 | 10.52 | 22.09 | 16.40 | 10.24 | 20.94 | 15.59 | 11.4 |

**Table 23**: Prediction Comparison for $1 - \alpha = 90\%, k = 1, h = 6$

The expected pattern is easily shown here; increasing values of $H$ increases the width, and increasing values of $k$ decreases the width.

### 4.2.2 LSTM model

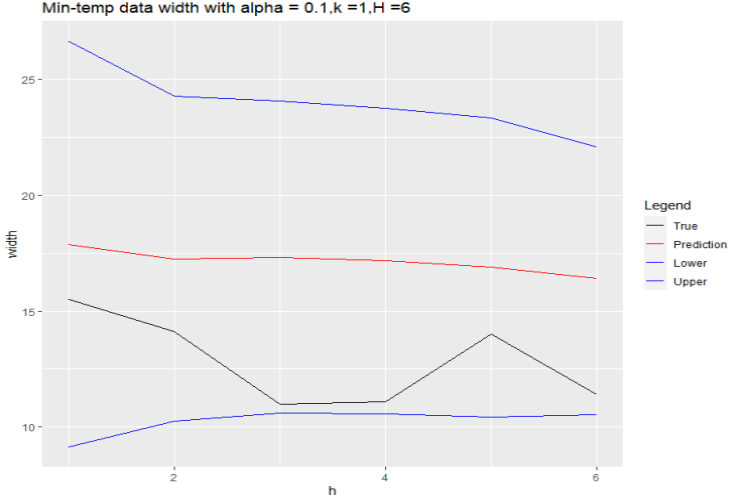 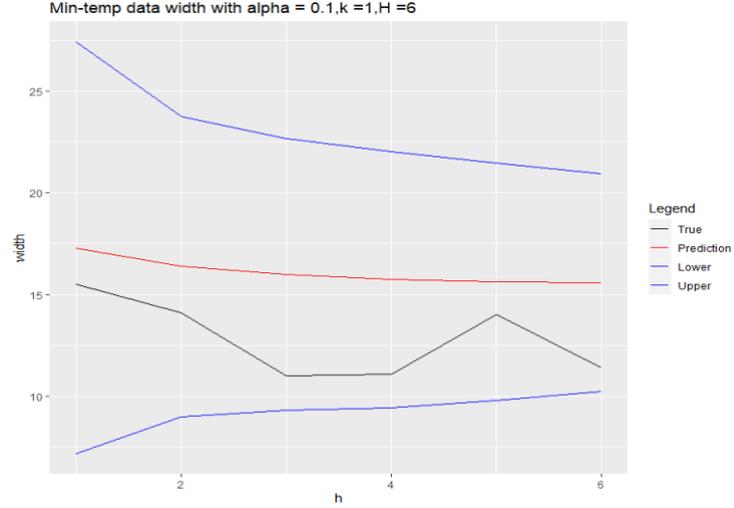

**Figure 9(a):** ARIMA JPR Min-temp data for $K = 1, H = 6, \alpha = 0.1$

**Figure 9(b):** LSTM JPR Min-temp data for $K = 1, H = 6, \alpha = 0.1$

The next model is an LSTM based neural net with a 1-D convolutional layer. The use of LSTM based Neural Network is expected to make the width smaller whilst computing the JPR with the desired coverage.

| Width, $1 - \alpha = 90\%$ | $H = 6$ | $H = 12$ | $H = 18$ | $H = 24$ |
|---|---|---|---|---|
| 1-FWE JPR | 13.57008 | 16.52786 | 20.24374 | 21.07529 |
| 2-FWE JPR | 9.54059 | 12.66316 | 16.19251 | 17.14191 |
| 3-FWE JPR | 6.806471 | 10.072410 | 13.517448 | 14.948179 |

**Table 24**: Prediction Width for $1 - \alpha = 90\%$

| Width, $1 - \alpha = 80\%$ | $H = 6$ | $H = 12$ | $H = 18$ | $H = 24$ |
|---|---|---|---|---|
| 1-FWE JPR | 11.58778 | 14.33384 | 17.59271 | 18.90147 |
| 2-FWE JPR | 7.856042 | 10.947055 | 14.428793 | 15.338635 |
| 3-FWE JPR | 5.663515 | 8.775633 | 12.221961 | 13.303556 |

**Table 25**: Prediction Width for $1 - \alpha = 80\%$

| Width, $1 - \alpha = 70\%$ | $H = 6$ | $H = 12$ | $H = 18$ | $H = 24$ |
|---|---|---|---|---|
| 1-FWE JPR | 10.19173 | 13.02902 | 16.12930 | 17.76648 |
| 2-FWE JPR | 6.985523 | 9.868055 | 13.271048 | 14.207637 |
| 3-FWE JPR | 5.069375 | 7.974999 | 11.172423 | 12.274985 |

**Table 26**: Prediction Width for $1 - \alpha = 70\%$

The observations follow the same pattern. However, there is an interesting pattern when compared with ARIMA's observations. The widths decreased for $H = 18$ when moved from LSTM to ARIMA but increased otherwise. Whilst it is unexpected, the difference is not significant considering the widths are computed only for one sequence; an average over all the sequences might give a better idea. Furthermore, different ARIMA models (in terms of parameters) were selected for all bootstraps, but the LSTM weights and architecture remained the same for all bootstraps. This is where the importance of good bootstrap sampling becomes apparent, as it enables using the same model weights and yield a good result.

Another important pattern about the computed JPRs for Temp data is that the width has decreased with $h$. This is unusual as width usually increases with $h$. This might be a hint that the assumptions made at the beginning are not holding true. The pattern is seen in both predictors' JPR. It could also be due to the nature of the data. In any case, further analysis is required to find the reason behind this pattern. It should be noted that forecasting at a different point may change the result.

## 4.3 Sunspot Dataset

The last dataset for which JPR is computed is the Sunspot Dataset. The dataset contains Monthly numbers of sunspots collected from the World Data Centre (SIDC). After converting to tsibble format, the dataset contains 3177 monthly observations collected from January 1749 to September 2013. The dataset is a part of R datasets and can be loaded and used by entering 'sunspot.month'. The sunspots are magnetic disks on the surface of the sun which appear as dark spots. They interfere with the propagation of radio waves and are important to be predicted for telecommunication companies to avoid system failures. The graphs below show the variation and decomposition of the series. The first graphs suggest a cycle every few years; the sunspot cycle typically ranges from 9 to 14 years. It doesn't have a fixed length and, therefore, can not be considered a seasonal component of the series. The second graph has decomposed the series with year seasonality.

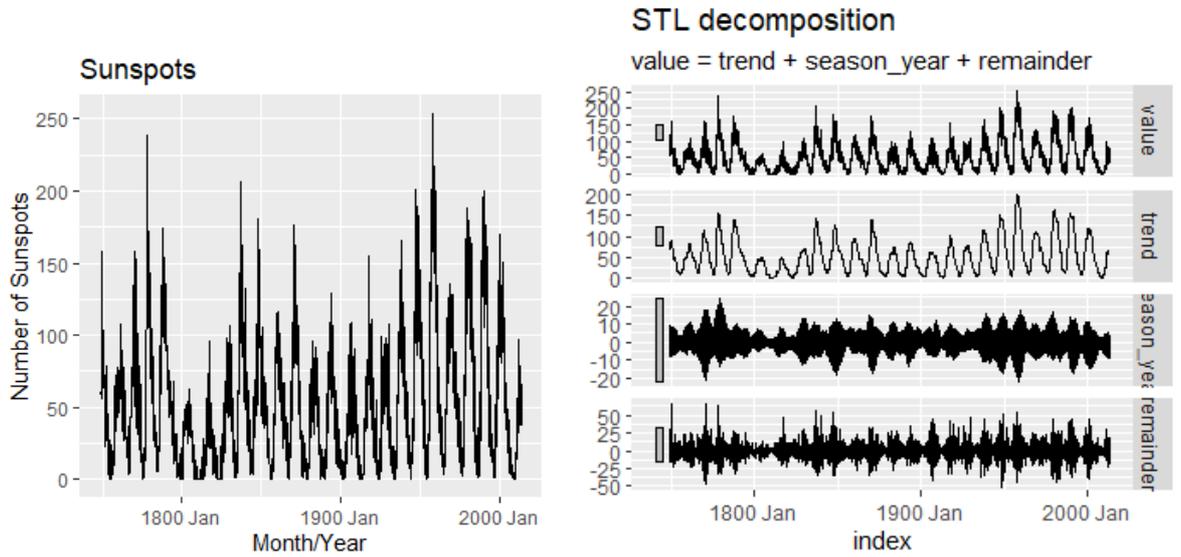

**Figure 10(a)** Plot for Sunspots Dataset

**Figure 10(b)** STL decomposition for Sunspots

Like the previous dataset, the geometric width is calculated for different $H$ and $k$ for one training set. Since the dataset is monthly, it makes more sense to predict for small $H$, but similar values of $H$ are compared regardless.

### 4.3.1 LSTM model

| Width, $1 - \alpha = 90\%$ | $H = 6$ | $H = 12$ | $H = 18$ | $H = 24$ |
|---|---|---|---|---|
| 1-FWE JPR | 87.48786 | 111.37350 | 129.95763 | 143.67568 |
| 2-FWE JPR | 61.23715 | 84.81054 | 103.83169 | 118.41853 |
| 3-FWE JPR | 46.19988 | 71.71352 | 93.30868 | 107.79830 |

**Table 27**: Prediction Width for $1 - \alpha = 90\%$

| Width, $1 - \alpha = 80\%$ | $H = 6$ | $H = 12$ | $H = 18$ | $H = 24$ |
|---|---|---|---|---|
| 1-FWE JPR | 68.84190 | 90.10743 | 110.55548 | 123.41896 |
| 2-FWE JPR | 50.10523 | 71.73123 | 91.07172 | 103.50447 |
| 3-FWE JPR | 36.40215 | 61.84014 | 79.62525 | 92.96417 |

**Table 28:** Prediction Width for $1 - \alpha = 80\%$

| Width, $1 - \alpha = 70\%$ | $H = 6$ | $H = 12$ | $H = 18$ | $H = 24$ |
|---|---|---|---|---|
| 1-FWE JPR | 59.10034 | 79.40858 | 99.00553 | 110.07776 |
| 2-FWE JPR | 42.19545 | 63.36357 | 80.97839 | 91.76209 |
| 3-FWE JPR | 31.02544 | 53.11339 | 70.26676 | 81.63254 |

**Table 29:** Prediction Width for $1 - \alpha = 70\%$

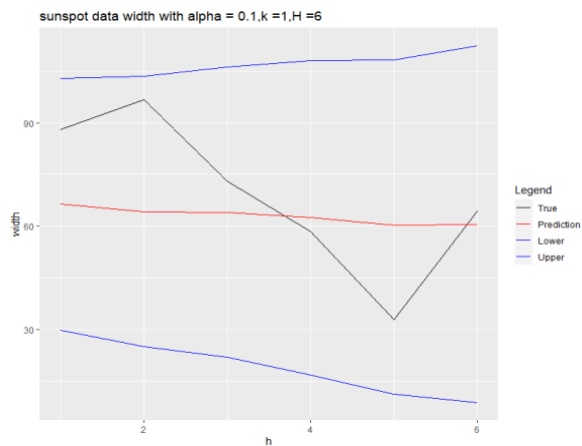

**Figure 11:** Sunspot data JPR for $K = 1, H = 6, \alpha = 0.1$

| $h$ | lower | upper | prediction | True value |
|---|---|---|---|---|
| 1 | 29.79 | 102.97 | 66.38 | 88 |
| 2 | 24.94 | 103.57 | 64.26 | 96.7 |
| 3 | 21.87 | 106.19 | 64.03 | 73 |
| 4 | 16.89 | 107.99 | 62.44 | 58.3 |
| 5 | 11.25 | 108.25 | 60.25 | 32.9 |
| 6 | 8.79 | 112.31 | 60.55 | 64.3 |

**Table 30:** JPR for $1 - \alpha = 90\%, k = 1, H = 6$

The results obtained reiterate the same conclusions drawn from the other dataset. However, for the sunspot dataset, the width of JPR increases with increasing $h$, as shown in figure 11, which is an expected outcome.

The huge prediction width with the sunspot data is largely due to a bigger range than the other datasets. It is also a dataset where careful bootstrapping is required as no possible value in the bootstrapped series can be negative. These factors can negatively influence the multiplier. Another thing to observe is that whilst the chosen method, which the trained network perform exceedingly well on one step predictions, it fails to generalise beyond a point by using the current prediction to find the next step's prediction and so on in a rolling window. This problem can be mitigated by an encoder-decoder LSTM architecture, which could potentially provide tighter bounds for predictions beyond the first step.

## 4.4 Conclusions

This chapter discussed the results obtained after applying different JPR methods, mainly focussing on k-FWE JPR. The results were obtained after assuming that the bootstrapped prediction standard errors are approximately the same as prediction standard errors, saving a lot of computation time. In order to reduce computation, only 100 bootstrap samples were used to obtain empirical coverages and empirical widths. Doing this provides a crude approximation of results, but it still provides a good comparison with other JPR methods and the effect of hyperparameters. In a nutshell, large values of $H$ increase widths or volume of JPR, whereas large values of $k$ decrease widths. Larger values of significance levels also decrease widths. Joint Marginals with Bonferroni correction causes loss of information and higher than desired coverage. Modified Scheffé JPR causes smaller coverages than NP heuristic JPR and k-FWE JPR (results in [Appendix](Appendix)). NP heuristic provides lesser widths and is sensitive to the number of bootstrap samples. Using stronger predictors results in tighter JPRs.

# 5   Professional Issues

In this section, I would talk about two professional issues: the reliability of results produced by a machine learning algorithm and the falsification of results. Most of the prediction results reported do not highlight their relevance. They provide no indication of the extent of trust one should have before using a prediction. Especially when different methods/ models are considered, it is better to have an idea of how reliable a particular model is before it could be deployed. As data scientists/ machine learning engineers, it is our responsibility to ensure our work is accurate and deployment in the real world would not cause harm of any kind. I chose this topic because the relevancy of results in machine learning is often ignored. However, it is of high importance to have relevancy in our work.

Relevancy can be showed by using methods that can show confidence in obtained results. In statistical applications that revolve around parameter estimation, it is typically done by using confidence intervals. There are a few methods, which can show relevance in results when the concerned problem is about prediction. One such method is bootstrapping, which typically uses sampling with replacement and makes several thousand proxy datasets and uses the variation in results to provide a prediction interval. If the obtained prediction interval is small and we have a coverage guarantee, we can be very confident in our results and can deem our findings reliable and be confident of deployment.

One other method is to use a framework called conformal predictors, which provides coverage guarantees and is much more suitable than bootstrapping for machine learning tasks. It also provides a convenient way to provide confidence for every prediction. There are several more methods like cross-validation and Jack-Knife+, which can help measure relevancy. This way of obtaining prediction intervals can be very helpful; if the calculated prediction intervals are very wide, it will make sense not to trust our prediction too much. On the other hand, if the prediction intervals are narrow, we can be sure that using the provided prediction in an application would yield a good result. This way, prediction intervals can remark on the accuracy of a model or method.

In this dissertation, I have discussed Joint Prediction Regions for time series, which is another attempt in the direction of showing reliability. It works with time series sequences and can be helpful in many tasks where a sequence needs to be predicted with high confidence. Some examples include stock market price prediction or blood glucose prediction. In some applications, taking the uncertainty can benefit while designing a system; if a system needs a conservative approach, upper/lower bound of prediction intervals can be used to ensure safe deployment.

To reiterate, it is necessary for many applications to know the confidence of results, especially in Health Sector and Automation Sector. To provide an example, there is a huge cost involved if the wrong medical treatment is provided based on some output of Machine Learning. Similarly, in self-driving cars, it is important to compute the relevancy of the decision before taking it, as a decision with low relevancy can have serious ramifications.

Research misconducts can be majorly classified into three categories: fabrication, falsification, and plagiarism. For the second issue, I focus on the falsification aspect of misconduct. Since I wish to work professionally in research, I chose to discuss this topic in detail. Results can be falsified in many ways, and it is more common to show misleading findings than believed. According to this news article[18], 8% of researchers fabricated or falsified data. I chose the falsification aspect because I recently met a person who falsified their findings, and despite my continuous attempts to show that it is wrong, they proceeded with it. Doing this is not only unprofessional but is also unethical and can have serious ramifications. In data related disciplines, it commonly happens after data collection; however, it may also happen before or during data collection. Since it usually happens after data collection, it is likely to happen in data processing, result in representation or sometimes in the statistical analysis stage.

There can be many scenarios and many ways this falsification can be done. I will talk about a few of them.

*Data Fishing-* Data Fishing refers to the phenomenon of reporting findings that proves the hypothesis. One example might be running 100 tests and only reporting the few where the desired outcome was achieved. This is wrong as the desired outcome may very well be a result of chance.

*Data Trimming-* As the data can be vast and may contain different patterns, Data Trimming can be done to isolate the pattern of interest to falsify the findings. Due to data trimming, the reported findings may not apply in the real world.

*Sampling Bias-* Sampling bias typically occurs when the sample used to draw inference from a population is no longer representative of the population. This can be part of falsification before data collection. A person conducting the experiment could deliberately choose a biased dataset to prove a hypothesis or show overly optimistic or pessimistic results.

*Choosing Metric-* In machine learning, especially in prediction tasks, it is necessary to have a method to measure the performance of the model. These methods are called metrics, and there are several metrics, especially for regression problems. Some popular Metrics are Mean Squared Error, Mean Absolute Error, Mean Absolute Percentage Error. We encountered this problem while working on a dataset in the past; the same task, but different models may have different metrics preferring one over the other. In cases of comparison, it might be tempting to say just show one metric which gets our point across but can actually be misleading.

*Graphical Manipulation-* Graphical manipulation is probably the most common type of falsification. It is common to mislead using graphical methods; some very common examples include not showing the axis of graphs.

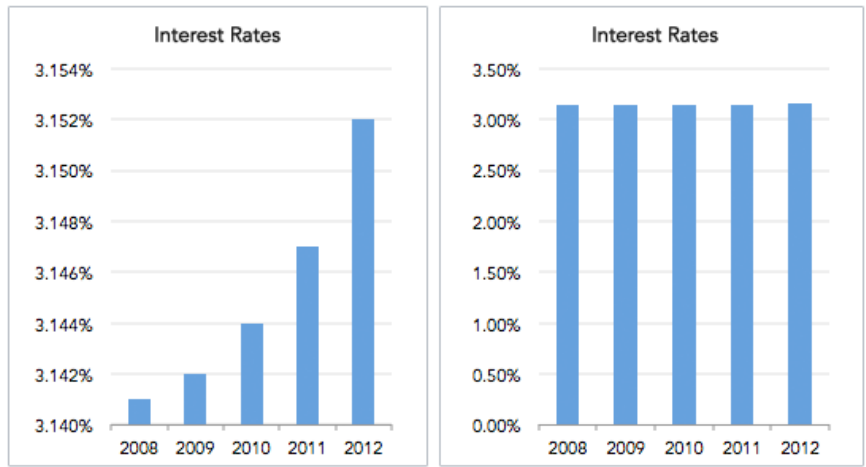

**Figure 12**: bar plot showing graphical simulation[18]

This figure shows two bar plots that show the same results, but the one on the left has a truncated y-axis, which changes everything. The graph on the left shows an increase in interest rates, but the graph on the right shows no significant difference. It can be tempting to remove the y-axis labels and mislead the findings.

I would like to end this discussion by reiterating the importance of providing unbiased and unfalsified findings. Many findings and proposals can be used in prominent sectors like the health sector, and employing methods obtained after falsification may cause significant losses.

# 6 Self-Assessment

Working on a dissertation has been a new experience for me. During my undergraduate studies, my major project was a group project and whilst it was a major project, the nature of it was very different from a dissertation like this one. This project has helped me develop many skills, explored many areas. It has taught me project planning, time management, improved my report writing skills.

I chose this project because it seemed to be a challenging yet fascinating project. It appeared to be a project which could help me explore those areas which were not covered in the module. I saw an opportunity to combine what I learnt from many modules into a single project and learn more. And I feel happy that I chose this project. It helped me understand advanced statistical techniques, apply Deep Learning models and methods discussed in the Online Machine Learning module to datasets. This project also helped me learn TensorFlow and use R for time series analysis. This project has fulfilled most of my reasons to choose this project.

Reflecting on my journey on this project, I started Wolf et al.'s paper with enthusiasm and quickly found out that I need to do more background reading than I have already done. I am thankful to have a supportive supervisor in Dr Nicola Paoletti. He guided me to helpful resources and helped me through this journey. He provided some very useful resources at the very beginning. I perused those resources about bootstrapping and time series. After that, I felt much more confident to start reading the paper. During our second meeting, I decided to use R for coding the JPR. Whilst an unusual choice, I used R as it helps me think mathematically when I code. Luckily, many resources using R were available to me for time series analysis and many statistical tests. I coded a block bootstrap method in R and completed my first set of background readings. This was when I started becoming over-ambitious; I started to compare the bootstrapping method with conformal predictors and relayed my desire to my supervisor, Nicola, who provided me with some research papers around the topic. I started re-reading my course material on Conformal Predictors, and this time, I developed a better understanding of the topic. I also started to read a little about the Jack-Knife method. I started thinking of ways to use Conformal Predictors to construct a JPR and compare it with existing methods[11]. At times I felt I was close, but later on finding a flaw in my approach. During this time, I fell ill, and sometime later, after my first vaccination, I suffered side effects, and it all took about two weeks to heal.

By this time, I realised the time was coming to an end. I restudied the research paper and quickly coded my first JPR implementation. This was when I realised an unsolved issue of finding prediction standard errors. I thought about it for a while and proposed a method to use double bootstrapping to estimate prediction standard errors and explained it to my supervisor and once I got his advice on the method. I used the same. At the same time, I started an online course on the Coursera platform about Time Series and Sequences deep learning networks[5], where I learnt TensorFlow to use for my project. I first started computing JPR for a stock market dataset[4]. It turned out that the nature of the stock market dataset is highly dynamic, and the dataset at hand is likely to perform poorly. I also learnt about the Efficient market hypothesis. However, the biggest challenge I found was computation time; using double bootstrapping with 100 samples took a considerate amount of time, and this time was bound to increase as 100 JPR's needed to be constructed to obtain empirical coverage. It was obvious that using Deep Learning models would only cause a problem. That's when I made some assumptions to decrease computation time and created a synthetic dataset to be sure that the assumption holds. This is how I computed empirical coverages and widths. After so many challenges, I moved to a real dataset, computed JPR and compared some predictors and other methods for constructing JPR. This has been the story with highs and lows, with challenges and excitements on this project.

    Towards the end of this project, I am happy to be able to work on the first two extensions but disappointed to not work on the third one, which looked exciting. Ironically, this project keeps having open ends, and new ideas keep coming to mind. Some of those ideas leave a regret; I hoped to have used Conformal Predictors in the project. I wanted to do Monte Carlo simulations and apply different bootstrap techniques. I also wanted to find a computationally efficient method for finding prediction standard errors. In the end, I conclude my assessment by saying, I learnt many things during this project, and it has developed me to read, understand high-level research papers, develop my theory, but I wished I had done more.

# 7 Bibliography

[1] Wolf, Michael, and Dan Wunderli. "Bootstrap joint prediction regions." Journal of Time Series Analysis 36.3 (2015): 352-376.

[2] Chapter 6, Lecture notes of 36-402, CMU Undergraduate Advanced Data Analysis, available at https://www.stat.cmu.edu/~cshalizi/ADAfaEPoV/ADAfaEPoV.pdf

[3] Zhang, A., Lipton, Z. C., Li, M., & Smola, A. J. (2019). Dive into deep learning, chapters 8 and 9, available at https://d2l.ai/index.html

[4] Stock Market Predictions with LSTM in Python, available at https://www.datacamp.com/community/tutorials/lstm-python-stock-market

[5] Sequence time series and prediction course Sequences, Time Series and Prediction - Home | Coursera

[6] Forecasting Principles and practice Forecasting: Principles and Practice (3rd ed) (otexts.com)

[7] Time Series Analysis and Its Applications With R Examples — 4th Edition Time Series Analysis and Its Applications: With R Examples - tsa4 (pitt.edu)

[8] Jordà Ò, Marcellino MG. 2010. Path-forecast evaluation. Journal of Applied Econometrics 25: 635–662.

[9] Staszewska-Bystrova A. 2011. Bootstrap prediction bands for forecast paths from vector autoregressive models. Journal of Forecasting 30(8): 721–735.

[10] Staszewska-Bystrova A. 2013. Modified Scheffé's prediction bands. Jahrbücher für Nationalökonomie und Statistik 233(5+6): 680–690.

[11] Lahiri, S. N. (2003). Resampling Methods for Dependent Data. New York: Springer-Verlag. 502, 519

[12] De Gooijer JG, Hyndman RJ. 2006. 25 years of time series forecasting. International Journal of Forecasting 22: 443–473.

[13] Conformal Prediction interval for dynamic time series. [2010.09107] Conformal prediction interval for dynamic time-series (arxiv.org)

[14] Modified scheffe Prediction Bands Modified Scheffé's Prediction Bands (degruyter.com)

[15] Classical F-test F-Test (hslu.ch)

[16] Jason Brownlee github jbrownlee/Datasets: Machine learning datasets used in tutorials on MachineLearningMastery.com (github.com)

[17] Efron, Bradley (1979). "Bootstrap Methods: Another Look at the Jackknife." Annals of Statistics, 7: 1–26. URL http://projecteuclid.org/euclid.aos/1176344552. doi:10.1214/aos/1176344552. 130, 152

[18] Misleading Data Visualization Examples (datapine.com)

[19] Box-Pierce test (sylwiagrudkowska.github.io)

[20] Ljung-Box test (sylwiagrudkowska.github.io)

# 8 How to Use My Project

To use this project's code, R and preferably R studio must be installed as most of the code is written in R. Two packages need to be installed before running the code: Kit library package, which provided 'topn' function used for speedy partial sort; and Fpp3 library package, which is a library written for Forecasting Principles and Practice book. The Fpp3 package is used to load and install many other packages required for modelling, like tibble, fable, and feasts. The packages can be downloaded from R by entering the following command:

install.packages(c('kit','fpp3'))

There is some piece of code, which is written in Python. The intermediate result files from Python and some R scripts are uploaded on Drive. A link is provided in the readme file to download the intermediate file and all the code. This can be used to import the intermediate results in R to get the final results. Python code is used for training a neural network and making forecasts. If running the python code or full R code is desired instead of using the intermediate file, the code is provided. Python code can be run on any machine with Python installed alongside TensorFlow, NumPy, and Pandas library.

To reproduce the result from a section, the relevant section folder contains standalone files which can be run to reproduce any result. The code is structured so that each file has all the needed functions, which makes it easy to analyse and understand the components of a code. All the code files structured in an orderly fashion and can also be downloaded using the following link:

https://drive.google.com/drive/folders/1qVbMBgubTRvwSuuVoTs7cRBGruC6LBhp?usp=sharing

# 9 Appendix

This chapter includes secondary results and all the code files used in this project.

## 9.1 NP Heuristic

This section shows the results obtained after computing NP heuristic JPR for the synthetic dataset.

### 9.1.1 Empirical Coverages and Empirical widths

This subsection shows the empirical coverages and empirical widths obtained for NP heuristic JPR. The results show under coverage, which is primarily attributed to a suboptimal number of bootstrap samples used, i.e. 100.

| 100 samples | $H = 6$ | $H = 12$ | $H = 18$ | $H = 24$ |
|---|---|---|---|---|
| $1 - \alpha = 90\%$ | 81 | 68 | 59 | 50 |
| $1 - \alpha = 80\%$ | 70 | 60 | 51 | 41 |
| $1 - \alpha = 70\%$ | 62 | 54 | 46 | 31 |

**Table 31**: Empirical Coverages for NP heuristic JPR

| 100 samples | $H = 6$ | $H = 12$ | $H = 18$ | $H = 24$ |
|---|---|---|---|---|
| $1 - \alpha = 90\%$ | 12.07527 | 12.35012 | 12.47038 | 12.56067 |
| $1 - \alpha = 80\%$ | 11.40547 | 11.82408 | 12.02736 | 12.14685 |
| $1 - \alpha = 70\%$ | 10.76646 | 11.30243 | 11.56238 | 11.70909 |

**Table 32**: Empirical Widths for NP heuristic JPR

### 9.1.2 Width Comparison with different sizes of bootstrap sampling

The tables below show a width comparison with varying sizes of bootstrap sampling. Increasing the number of bootstrap samples results in higher width, and the tables below show that point. It is expected that sufficiently large bootstrap samples provide a better estimation. Since the previous subsection has demonstrated that the empirical coverage obtained from 100 bootstrap samples causes under coverage. It can be inferred from comparing the widths of different sizes of bootstrap samples that better empirical coverage can be obtained by more bootstrapping. We wish to reiterate that coverage results provided in the previous subsection still offer a good comparison with other methods and hint at k-FWE as being a more robust JPR method.

| 100 samples      | $H = 6$  | $H = 12$ | $H = 18$ | $H = 24$ |
|------------------|----------|----------|----------|----------|
| $1 - \alpha = 90\%$ | 11.04928 | 12.26697 | 12.98398 | 13.22536 |
| $1 - \alpha = 80\%$ | 10.34406 | 11.72682 | 12.45595 | 12.70891 |
| $1 - \alpha = 70\%$ | 10.03182 | 11.01840 | 12.21616 | 12.19899 |

**Table 33**: Calculated Widths for 100 bootstrap samples

| 1000 samples     | $H = 6$  | $H = 12$ | $H = 18$ | $H = 24$ |
|------------------|----------|----------|----------|----------|
| $1 - \alpha = 90\%$ | 15.35395 | 15.51012 | 16.20209 | 16.07394 |
| $1 - \alpha = 80\%$ | 14.25186 | 14.85779 | 15.37549 | 15.31328 |
| $1 - \alpha = 70\%$ | 13.44665 | 14.23607 | 15.01044 | 14.91798 |

**Table 34**: Calculated Widths for 1000 bootstrap samples

| 10,000 samples   | $H = 6$  | $H = 12$ | $H = 18$ | $H = 24$ |
|------------------|----------|----------|----------|----------|
| $1 - \alpha = 90\%$ | 16.62640 | 17.34424 | 18.81650 | 19.08783 |
| $1 - \alpha = 80\%$ | 15.44308 | 16.66712 | 17.89279 | 18.48866 |
| $1 - \alpha = 70\%$ | 13.93647 | 16.12108 | 17.40507 | 18.15059 |

**Table 35**: Calculated Widths for 10,000 bootstrap samples

The above width comparison shows an increase in width with larger bootstrap samples and the possibility of obtaining desired coverage with more bootstrap samples. It must also be noted that the NP heuristic method did not require any assumption and just requires a good bootstrapping method to compute its prediction region.

## 9.2 Modified Scheffé JPR

This section shows the results obtained after computing Modified Scheffé JPR for the synthetic dataset. Just like the coverages obtained from the NP heuristic method, there is under coverage. However, the result shows serious under coverage. The results are aligned with the expectation to an extent as under coverage is expected with this JPR method. We can still draw comparisons with other JPR methods. Modified Scheffé JPR has performed worse than NP heuristic and k-FWE JPR. However, we would conclude by saying that these results require further analysis, especially the bootstrapping method used to estimate the covariance matrix.

| 100 samples | $H = 6$ | $H = 12$ | $H = 18$ | $H = 24$ |
|---|---|---|---|---|
| $1 - \alpha = 90\%$ | 56 | 44 | 42 | 42 |
| $1 - \alpha = 80\%$ | 41 | 29 | 26 | 26 |
| $1 - \alpha = 70\%$ | 24 | 15 | 14 | 14 |

**Table 36**: Empirical Coverages for Modified Scheffé JPR

| 100 samples | $H = 6$ | $H = 12$ | $H = 18$ | $H = 24$ |
|---|---|---|---|---|
| $1 - \alpha = 90\%$ | 14.11119 | 15.90500 | 17.75534 | 19.55518 |
| $1 - \alpha = 80\%$ | 11.94075 | 13.83178 | 15.66095 | 17.40905 |
| $1 - \alpha = 70\%$ | 10.43228 | 12.38073 | 14.19052 | 15.89950 |

**Table 37**: Empirical Coverages for Modified Scheffé JPR

## 9.3 Joint Marginals LSTM Synthetic Data

This section shows the results of Joint Marginals computed for Synthetic Data. This set of results can be considered as failed results. Using bootstrapping is a popular and convenient method used for calculating prediction intervals. However, if the number of bootstrap samples is less, the results can fail miserably. This section shows how adversely affected the results can be if the derived results are highly dependent on bootstrapping. The tables below show the empirical coverages obtained on the synthetic dataset after applying Bonferroni Correction on Joint Marginals.

| 100 samples | $H = 6$ | $H = 12$ | $H = 18$ | $H = 24$ |
|---|---|---|---|---|
| $1 - \alpha = 90\%$ | 79 | 69 | 63 | 51 |
| $1 - \alpha = 80\%$ | 75 | 64 | 59 | 47 |
| $1 - \alpha = 70\%$ | 69 | 60 | 53 | 41 |

**Table 38**: Empirical Coverages for Joint Marginals BF Correction

| 100 samples | $H = 6$ | $H = 12$ | $H = 18$ | $H = 24$ |
|---|---|---|---|---|
| $1 - \alpha = 90\%$ | 11.68161 | 12.33869 | 12.60148 | 12.68899 |
| $1 - \alpha = 80\%$ | 10.68940 | 11.65221 | 12.09026 | 12.34794 |
| $1 - \alpha = 70\%$ | 9.99170 | 11.12743 | 11.39600 | 12.00361 |

**Table 39**: Empirical Widths for Joint Marginals BF Correction